\documentclass{article}

\usepackage{placeins}

\usepackage{subcaption}
\usepackage{arxiv}

\usepackage[utf8]{inputenc} 
\usepackage[T1]{fontenc}    
\usepackage{hyperref}       
\usepackage{url}            
\usepackage{booktabs}       
\usepackage{amsfonts}       
\usepackage{nicefrac}       
\usepackage{microtype}      
\usepackage{lipsum}		
\usepackage{graphicx}
\usepackage{doi}
\usepackage{CJKutf8}
\usepackage{multirow}

\title{A New Amharic Speech Emotion Dataset and Classification Benchmark}

\author{ \mbox{\href{https://orcid.org/0000-0003-0906-9605}{\includegraphics[scale=0.06]{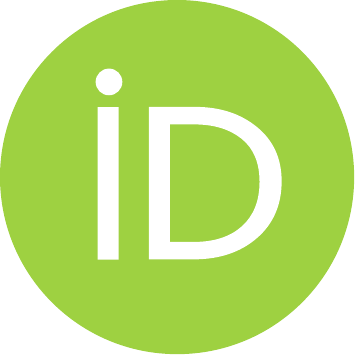}\strut}}\hspace{1mm}Ephrem Afele Retta \\
	School of Information Science and Technology\\
	Northwest University\\
	Xi’an 710127, China \\
	\texttt{afele@stumail.nwu.edu.cn} \\
	\And
	\mbox{\href{https://orcid.org/0000-0002-7182-9639}{\includegraphics[scale=0.06]{orcid.pdf}\strut}}\hspace{1mm}Eiad Almekhlafi \\
	School of Information Science and Technology\\
	Northwest University\\
	Xi’an 710127, China \\
	\texttt{ealmekhlafi@stumail.nwu.edu.cn} \\
	\And
	\mbox{\href{https://orcid.org/0000-0002-5549-5691}{\includegraphics[scale=0.06]{orcid.pdf}\strut}}\hspace{1mm}Richard Sutcliffe\thanks{Corresponding author} \\
	School of Information Science and Technology\\
	Northwest University\\
	Xi’an 710127, China \\
	School of Computer Science and Electronic Engineering \\ University of Essex \\
	Wivenhoe Park, Colchester CO4 3SQ, UK\\
	\texttt{rsutcl@nwu.edu.cn, rsutcl@essex.ac.uk} \\
	\And
	\mbox{\href{https://orcid.org/0000-0002-3106-669X}{\includegraphics[scale=0.06]{orcid.pdf}\strut}}\hspace{1mm}Mustafa Mhamed \\
	School of Information Science and Technology\\
	Northwest University\\
	Xi’an 710127, China \\
	\texttt{mustafamhamed@stumail.nwu.edu.cn} \\
	\And
	\mbox{\href{https://orcid.org/0000-0003-2064-5079}{\includegraphics[scale=0.06]{orcid.pdf}\strut}}\hspace{1mm}Haider Ali \\
	School of Information Science and Technology\\
	Northwest University\\
	Xi’an 710127, China \\
	\texttt{alihaider@stumail.nwu.edu.cn} \\
	\And
	\mbox{\href{https://orcid.org/0000-0002-0706-2103}{\includegraphics[scale=0.06]{orcid.pdf}\strut}}\hspace{1mm}Jun Feng\thanks{Corresponding author}\\
	School of Information Science and Technology \\
	Northwest University\\
	Xi’an 710127, China \\
	\texttt{fengjun@nwu.edu.cn} \\
}

\date{}

\renewcommand{\headeright}{E. A. Retta et al.}
\renewcommand{\undertitle}{}
\renewcommand{\shorttitle}{Amharic Speech Emotion Dataset}

\hypersetup{
pdftitle={A template for the arxiv style},
pdfsubject={q-bio.NC, q-bio.QM},
pdfauthor={David S.~Hippocampus, Elias D.~Striatum},
pdfkeywords={First keyword, Second keyword, More},
}

\begin{document}
\maketitle

\begin{abstract}
In this paper we present the Amharic Speech Emotion Dataset (ASED), which covers four dialects (Gojjam, Wollo, Shewa and Gonder) and five different emotions (neutral, fearful, happy, sad and angry). We believe it is the first Speech Emotion Recognition (SER) dataset for the Amharic language. 65 volunteer participants, all native speakers of Amharic, recorded 2,474 sound samples, two to four seconds in length. Eight judges (two for each dialect) assigned emotions to the samples with high agreement level (Fleiss kappa = 0.8). The resulting dataset is freely available for download. Next, we developed a four-layer variant of the well-known VGG model which we call VGGb.
Three experiments were then carried out using VGGb for SER, using ASED. First, we investigated whether Mel-spectrogram features or Mel-frequency Cepstral coefficient (MFCC) features work best for Amharic. This was done by training two VGGb SER models on ASED, one using Mel-spectrograms and the other using MFCC. Four forms of training were tried, standard cross-validation, and three variants based on sentences, dialects and speaker groups. Thus, a sentence used for training would not be used for testing, and the same for a dialect and speaker group. The conclusion was that MFCC features are superior under all four training schemes. MFCC was therefore adopted for Experiment 2, where VGGb and three other existing models were compared on ASED: RESNet50, Alex-Net and LSTM. VGGb was found to have very good accuracy (90.73\%) as well as the fastest training time. In Experiment 3, the performance of VGGb was compared when trained on two existing SER datasets, RAVDESS (English) and EMO-DB (German) as well as on ASED (Amharic). Results are comparable across these languages, with ASED being the highest. This suggests that VGGb can be successfully applied to other languages. We hope that ASED will encourage researchers to explore the Amharic language and to experiment with other models for Amharic SER.
\end{abstract}


\section{Introduction}
Emotion plays a significant role in everyday human interactions \cite{kerkeni2019automatic}. It is essential to rational decision making and helps us match and understand others' feelings by 
conveying our own feelings and giving feedback to others. 
Emotion conveys considerable information about the mental state of an individual.
This has opened up a new research field called automatic emotion recognition.
Various modalities have been explored in prior studies to recognize emotional states such as facial expressions, speech, physiological signals, etc. \cite{kerkeni2019automatic}. Several inherent advantages make speech signals a good source for affective computing. For example, compared to many other biological signals (e.g., electrocardiograms), speech signals can usually be acquired more readily and economically. This is why many researchers are interested in Speech Emotion Recognition (SER).

SER is an important research area that has been active for more than two decades \cite{schuller2018speech}. The results of SER can already be seen in many application fields, including entertainment, computer games, audio monitoring, online learning, clinical research, polygraph tests, and call centers \cite{kwon2020cnn,kerkeni2019automatic,basu2016effects}.
Even though SER has many benefits, it is still a difficult task to perform with high accuracy \cite{kwon2020clstm}. One key problem is choosing the right features; an incorrect choice can lead to moderate performance \cite{issa2020speech}. Audio features are usually divided into two domains; time-domain features and frequency-domain features. The time-domain functions are elementary to extract and allow easy analysis of audio signals. 
In the case of small audio datasets, the frequency domain features will show deeper patterns, which may help distinguish the signal's basic emotion.
Frequency-domain features include spectrograms, Mel Frequency Cepstral Coefficients (MFCCs), spectral centroid, spectral roll-off, spectral entropy, and Chroma coefficients \cite{tomas2019speech}.
In this paper, a comprehensive analysis of each feature was performed during the exploratory data analysis.
However, for the purpose of this work, we limited ourselves to two principal features, Mel-spectrograms and MFCC.

In recent years, numerous speech datasets have been created to support deep learning for SER. The languages of these datasets include English, Chinese, Spanish, French, German and many others \cite{khalil2019speech}.
However, no SER dataset has been created for Amharic yet.
Amharic is the second-largest Semitic language in the world after Arabic and the national language of Ethiopia \cite{mossie2018social}. In terms of the number of speakers and the significance of its politics, history, and culture, it is one of the 55 most important languages in the world  \cite{mengistu2017text}.
However, despite this, Amharic and its dialects have very few language resources compared to other languages such as English. It is for this reason that we have created a new SER dataset for Amharic.

Unlike English, Amharic is a syllabic language; each character represents one syllable \cite{belay2020amharic}. The language uses a script derived from the Ge'ez alphabet \cite{belay2020amharic}. It has 33 main characters, and each consonant-vowel combination has seven forms. Compared with English, Amharic has some unique phonemes.
Gemination in Amharic is one of the most distinctive features of the speech's cadence, and it has great semantic and syntactic functional weight. In contrast to English, where the rhythm is mainly characterized by stress (loudness), the rhythm of Amharic is mainly characterized by longer and shorter syllables depending on the germination of consonants and certain characteristics of the phrase \cite{anberbir2011grapheme}.
Amharic gemination is either lexical or morphological. In the speech synthesis field, it is vital to introduce emotional features to give greater expression to a machine, so that it speaks more like a human.
Additionally, expressions and intonation for emotions and mental states differ markedly in their nature and significance from language to language.
 
In summary, therefore, the two main challenges for Amharic are the limited availability of datasets and the language's complex morphological characteristics \cite{mulugeta2012learning}. This paper addresses both points by introducing a new Amharic dataset suitable for training and testing SER systems, and by presenting experiments which show how frequency-domain features can be effectively used for Amharic SER.
                
In addition, we have developed an SER model called VGGb, based on the well-known VGG SER architecture, and used it to carry out three experiments. 
The first experiment was to choose an appropriate method from the Mel-spectrogram and Mel-frequency Cepstral Coefficient (MFCC) technologies for extracting features from recordings in our ASED dataset, using VGGb. Four train-test schemes were used. The first was standard cross-validation, withholding 10\% of training utterances for testing. The second trained on just five of the seven sentences in ASED, testing on the remaining two. The third trained only with utterances spoken in three of the four dialects in ASED, testing on utterances in the remaining dialect. The fourth trained using utterances by participants in two of the three speaker groups in ASED, testing on utterances in the remaining speaker group. Under all four schemes, MFCC was found to be the most effective in terms of accuracy and training time.

The second experiment was to compare the classification performance of VGGb and three other popular models using MFCC features. VGGb achieved a high accuracy (90.73\%) as well as having the shortest training time. In the third experiment, we evaluated VGGb on the English RAVDESS and German EMO-DB datasets. The results for English and German were comparable to those achieved for Amharic on the ASED dataset.

The contributions of this paper are as follows:
\begin{itemize}
\item We create for the very first time an SER dataset for Amharic, called ASED. There are 65 speakers and 2,474 recordings, 522 neutral, 510 fearful, 486 happy, 470 sad, and 486 angry.
\item All four Amharic dialects (Gojjam, Wollo, Shewa and Gonder) are included in the dataset.
\item Eight judges evaluate the recordings, and agreement between them is high (Fleiss kappa = 0.8). So the data is of high quality.
\item We analyse ASED with respect to nine other well-known SER datasets and show that it compares very well in terms of the number of participants, the amount of data produced, the method of quality control, the number of judges, and their agreement level.
\item We develop a high-performing variant of the VGG SER model which has just four CNN layers. We call this VGGb.
\item Using VGGb and ASED data, we compare Mel-spectrogram and Mel-frequency Cepstral Coefficient (MFCC) features and show experimentally in a SER task that MFCC leads to higher accuracy. We show that the superiority of MFCC is independent of training sentence, Amharic dialect and speaker group.
\item We apply VGGb and three other architectural models to the SER task, working with ASED data, and show that VGGb is very effective, and by far the fastest.
\item Using VGGb and working with MFCC, we train models to recognise five emotions in Amharic, English and German, using the ASED, RAVDESS, and EMO-DB datasets respectively. VGGb shows excellent performance and proves very effective for SER on our Amharic data.
\end{itemize}
The rest of this paper is organised as follows: Section 2 presents the ASED dataset, describing the rationale behind its design and the method by which it was created. A detailed comparison with nine other datasets is also included (four for English, one each for Chinese, German, Greek, Gujarati, Hindi). Section 3 discusses feature extraction for speech emotion recognition, briefly outlining Mel-spectrograms and Mel-frequency cepstral coefficients. Section 4 describes the VGGb architecture and settings used for our experiments.Sections 5-7 describe the experiments. Section 8 gives conclusions and next steps.
\section{ASED DATASET for AMHARIC}
\subsection{Existing SER datasets} \label{existing_ser_datasets}
Before presenting our proposed dataset, we briefly describe two well-known existing datasets which are used in our experiments. Table \ref{comparison-ASED-RAVDESS-EMO-DB} shows the main data.
The Ryerson Audio-Visual Database of Emotional Speech and Song (RAVDESS) \cite{livingstone2018ryerson} contains audio and video recordings of English sentences spoken by twelve males and twelve females in eight emotions: neutral, calm, happy, sad, angry, fearful, surprise, and disgust.
The total number of speech file utterances is 1,440.
Recordings are three seconds in length at a sampling rate of 48 kHz.
The Berlin Emo-DB dataset \cite{burkhardt2005database} contains audio recordings of German sentences made by five males and five females in seven diﬀerent emotions, neutral, fear, anger, happiness, sadness, disgust, and boredom. The speech material comprises about 535 sentences.  
Recordings are two to three seconds in length at a sampling rate of 16 kHz.
\begin{table}[ht]
        \centering
        \caption{Comparison between RAVDESS, Emo-DB and ASED datasets.}
           	 \begin{tabular}{cccc}
           		 \hline  
          Aspect	& RAVDESS & Emo-DB & ASED  \\\hline 
          Language & English & German & Amharic \\
          Number of recordings	&  1440 &535 &2474\\
          Number of sentences	& 2  &10  &27 \\
          Number of participants	&  24  &10  &65 \\
          Number of emotions     & 8 & 7  & 5\\
           		\hline 
           	\end{tabular}
           	\label{comparison-ASED-RAVDESS-EMO-DB}
           \end{table}  
\begin{table}[ht]
        \centering
        \caption{Some datasets for speech emotion recognition (Sp+Utt = Speakers and Utterances, Sampl = Sampling Frequency, Quant = Quantization, Env = Environment, Mod = Modalities, Sim = Simulated, Nat = Natural, Ind = Induced, Lab = Laboratory Environment, F/TV = Film/TV, A = Audio, V = Visual).}
           	 \begin{tabular}{ccccclccc}
           		 \hline  
Database & Language & Sp.+Utt.
& \begin{tabular}[c]{@{}l@{}}Sampl.\\ (kHz)\end{tabular}& Quant. & Emotions & Type& Env.&Mod.  \\\hline

AESDD  & Greek  &  5+500 & 44.1 &16 bits &  \begin{tabular}[c]{@{}l@{}}{\small anger, disgust}\\{\small fear, happiness}\\ {\small sadness.}\end{tabular} & Sim   & Lab & A \\
 
 ASED & Amharic & 65+2474 & 16 & 16 bits & \begin{tabular}[c]{@{}l@{}}{\small neutral, fearful}\\{\small happy, sad}\\ {\small angry.}\end{tabular} & Sim    & Lab  & A \\  
 
 CHEAVD & Mandarin & 238+2600 & 44.1 & 16 bits & \begin{tabular}[c]{@{}l@{}}{\small 26 emotions:}\\{\small surprise, happy }\\{\small sad, angry }\\  {\small fearful, neutral etc. }\end{tabular} &  Nat & F/TV & A/V \\
 
 EGSC &  Gujarati & 9+1296 & 44.1 & -- & \begin{tabular}[c]{@{}l@{}}{\small sadness, surprise}\\{\small anger, disgust }\\{\small  fear, happiness.}\end{tabular} & Sim & Lab & A \\
  
 EMO-DB & German & 10+535 & 16 & 16 bits & \begin{tabular}[c]{@{}l@{}}{\small anger, boredom}\\{\small disgust, fear}\\{\small sadness, neutral}\\ {\small happiness.}\end{tabular} & Sim & Lab & A \\

IEMOCAP & English &  10+1150 & 48 & 16 bits &  \begin{tabular}[c]{@{}l@{}}{\small neutral, happiness }\\{\small  anger, sadness}\\ {\small frustration.}\end{tabular} & Sim/Ind  & Lab & A/V \\

{\small IITKGP-SEHSC}  & Hindi &  10+12000 & 16 & 16 bits & \begin{tabular}[c]{@{}l@{}}{\small anger, disgust}\\ {\small fear, happy}\\{\small neutral, sad}\\ {\small sarcastic, surprise.}\end{tabular} & Sim  & Lab & A\\

 RAVDESS  & English &  24+4320 & 48 & 16 bits &  \begin{tabular}[c]{@{}l@{}}{\small neutral, calm}\\ {\small  happiness, sadness} \\ {\small anger, fear} \\ {\small surprise, disgust.}\end{tabular} & Sim & Lab& A/V \\
 
SAVEE & English &  4+480 & 44.1 & 16 bits & \begin{tabular}[c]{@{}l@{}}{\small anger, disgust}\\ {\small fear, happiness}\\{\small sadness, surprise}\\ {\small neutral, common.}\end{tabular} & Sim & Lab & A/V \\

TESS & English & 2+2800 & 24.4 & 16 bits &  \begin{tabular}[c]{@{}l@{}}{\small anger, disgust }\\ {\small neutral, fear}\\{\small happiness, sadness}\\ {\small pleasant, surprise.}\end{tabular} & Sim & Lab & A \\
                                             
           		\hline 
           	\end{tabular}
           
           	\label{ser-datasets}
           \end{table} 
  
\subsection{Design of ASED}	

       \begin{table}[ht]
            	\centering
            	\caption{ Examples of sentences expressing each of the five emotions.}
                	\includegraphics[width=130mm,scale=4]{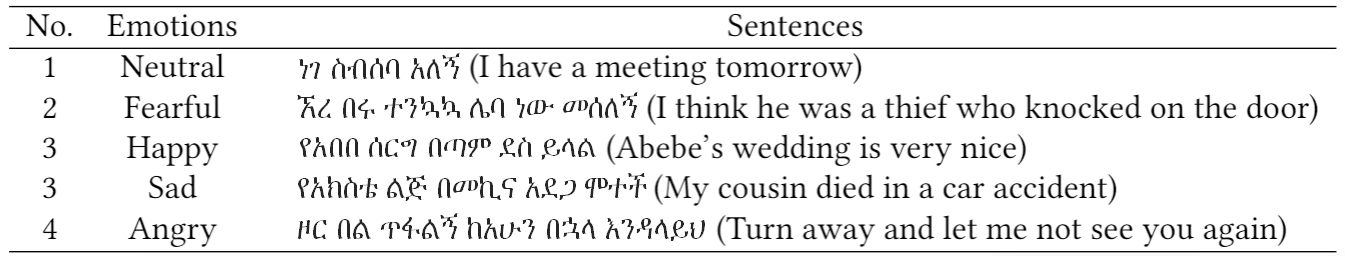}
                	
                	\label{sentences-expressing-emotions}
                \end{table} 
      
\textbf{Dialects:} Amharic is considered one of the most challenging languages to be utilized in speech emotion recognition systems because of its huge lexical variety and complicated morphology \cite{anberbir2011grapheme}.
There are four main types of Amharic dialect, namely Gojjam (Gojjamegna), Wollo (Wollogna), Shewa (Shewagna), and Gonder (Gonderegna) \cite{mengistu2017text}. We wished the proposed dataset to contain examples of all four dialects.

\textbf{Emotions:}
RAVDESS contains eight emotions (see above) while Emo-DB has seven. It was decided to adopt five emotions which are common to the other datasets.
So, relative to RAVDESS, calm, surprise and disgust are omitted, while 
relative to Emo-DB, disgust and boredom are omitted.
The use of a common subset allows direct SER comparisons to be made across languages.

\textbf{Test sentences:}
For each of the five emotions in ASED, five sentences expressing that emotion were composed in Amharic. For example, for Happy we have `Abebe's wedding is very nice' (Table \ref{sentences-expressing-emotions}), a sentence which expresses a happy concept. Similarly, `My cousin died in a car accident' is a Sad sentence. Furthermore the dataset contains two common sentences which express no strong emotion, e.g. `My sister is coming by plane'. The total number of sentences in the dataset is thus 27, 5 × 5 emotion-specific sentences plus two common sentences.

\textbf{Approach to generating emotion:}
Three methods can be used to collect the recordings in emotion recognition datasets: Simulated, Induced, and Natural \cite{khalil2019speech}. For simulated emotions, participants are asked to read a sentence while expressing a stated emotion. So if the sentence was `My sister is coming by plane', one participant could be asked to read it in an angry way, while another read it in a sad way. In the Induced approach, the participant is made to feel the required emotion before reading the sentence. In the Natural approach, a recording must be made when a speaker happens to be feeling a particular emotion. The Simulated approach is the most practical to implement, and nearly 60\% of speech datasets are collected using this method \cite{khalil2019speech}. For this reason, the Simulated approach was also adopted for ASED data collection.
\begin{table}[ht]
  \centering
        \caption{Number of utterances per speaker in the ASED database.}
\begin{tabular}{ccccccc} 
\hline 
Id & Emotion & \multicolumn{4}{c}{Amharic dialect}  & Number of                                    \\ \cline{3-6}
                    &                                    & Gojjam  & Wollo  & Shewa  & Gonder  &     recordings                                    \\ \hline
1                   & Neutral                            & 104      & 208    & 140    & 70      & 522                                     \\
2                   & Fearful                               & 59      & 170    & 200    & 81       & 510                                     \\
3                   & Happy                              & 68      & 150     & 188    & 80      & 486                                     \\
4                   & Sad                                & 70      & 135     & 135      & 130       & 470                                     \\
5                   & Angry                              & 111      & 120    & 140    & 115     & 486                                     \\
\multicolumn{6}{l}{Total}                                                                 & 2,474    \\ \hline                               
\end{tabular}
\label{no-utterances-per-speaker}
\end{table}
        
        \begin{figure}[h]
        	\centering
        	\includegraphics[width=63mm,scale=1.5]{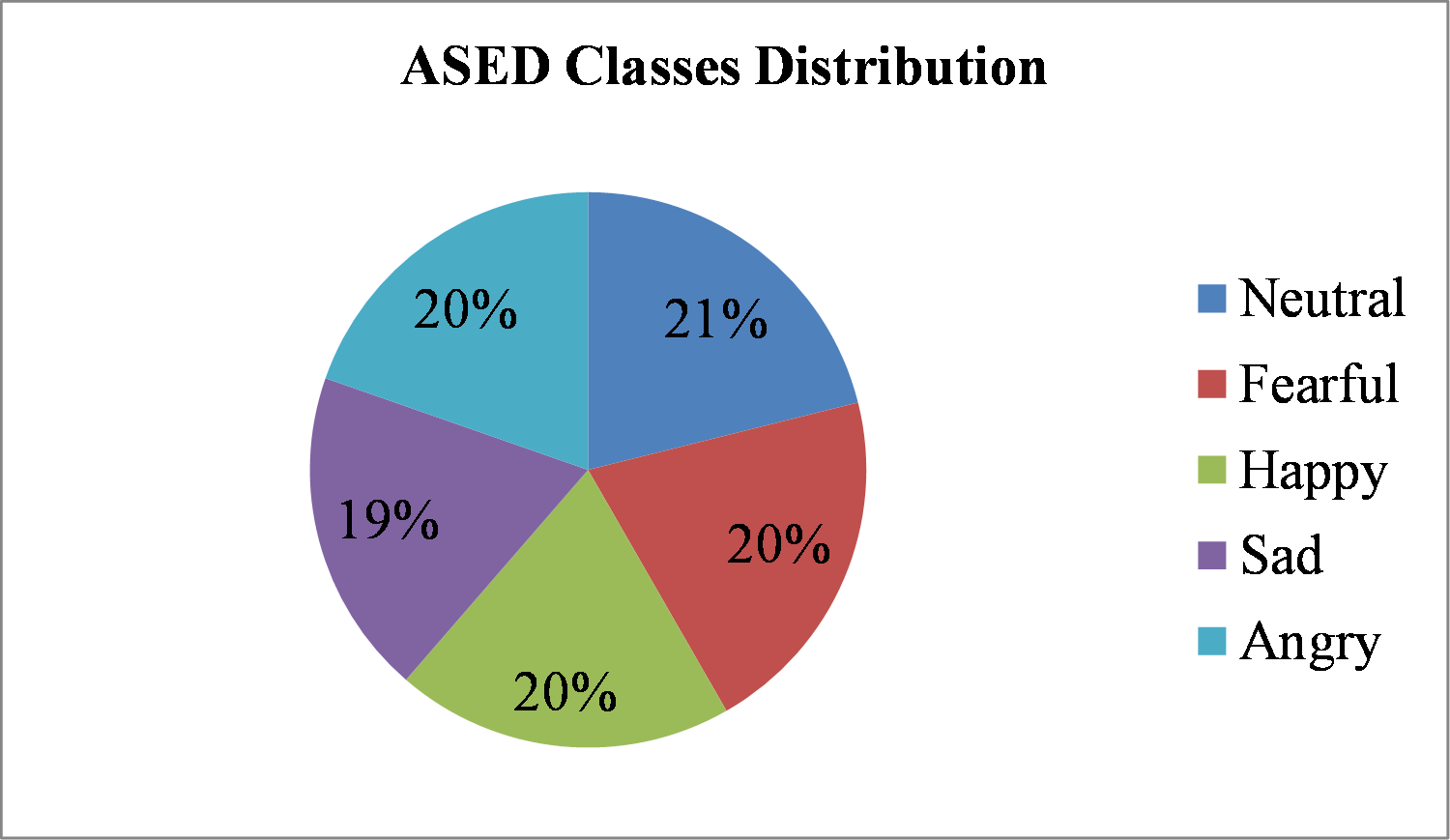}
        	\caption{ASED dataset distribution across all emotion classes.}
        	\label{ased-distribution}
        \end{figure}
    
\subsection{Creation of ASED}\label{creation-of-ased}
\textbf{Judges:}

All judges were Ethiopian postgraduate students of Computer Science or Management at universities in Xi'an, China. All were native speakers of one Amharic dialect. Two judges were from Xidian University
\begin{CJK*}{UTF8}{gbsn}
(西安电子科技大学)
\end{CJK*}
and spoke Shewa and Gonder dialects respectively; two were from Chang'an University
\begin{CJK*}{UTF8}{gbsn}
(长安大学)
\end{CJK*}
(Wollo, Gojjam); one was from Xi'an Shiyou University
\begin{CJK*}{UTF8}{gbsn}
(西安石油大学),
\end{CJK*}
(Shewa); two were from Xi'an Jiaotong University
\begin{CJK*}{UTF8}{gbsn}
(西安交通大学),
\end{CJK*}
(Gonder, Gojjam); finally, one was from Northwest University
\begin{CJK*}{UTF8}{gbsn}
(西北大学),
\end{CJK*}
(Wollo). Judges were responsible for the quality control of the dataset (see below).

\textbf{Participants:}
There were three classes of participant, undergraduate students, postgraduate students and business people. The undergraduate and postgraduate students came from Ethiopia to China in order to study. The business people came to China for professional reasons concerning their work. In order to take part, participants had to be native speakers of one of the four Amharic dialects, and they had to be capable of speaking in different emotions, according to the opinions of the judges, following some initial tests.
During the selection process, the judges assigned each participant to one of three groups, depending on their expertise in the task: Professional, Semi-professional and Amateur (Table \ref{Expert speakers}). Participants who were clearly experienced at acting and were excellent at expressing the different emotions, in the opinion of the judges, were assigned to the Professional group. Those who were judged very good at expressing the emotions were assigned to the Semi-professional group. Finally, participants who were not experienced at acting but were nevertheless judged good enough to participate in the task were assigned to the Amateur group.
In total there were  65 participants (25 female, 40 male), aged from 20 to 40 years.
 \begin{table}[ht]
        \centering
        \caption{Classification of 65 participants in ASED dataset.}
           	 \begin{tabular}{cc}
           		 \hline
           	Category & Values \\\hline
            Occupation       &  Postgraduate (21), Undergraduate (21), Business person (23)  \\\hline 
            Expertise      &  Professional (20), Semi-professional (26), Amateur (19)  \\
             
           		\hline 
           	\end{tabular}
           	\label{Expert speakers}
           \end{table} 

\textbf{Recording:}
In order to record the speech, we used six Huawei nova4 mobile phones, on which an Android-based speech recording software app \cite{anberbir2011grapheme} had been installed. Mobile phones were used because professional audio equipment was not available to us. The software was set up to capture the speech utterances utilizing a 16 kHz sampling rate at 16 bits, resulting in a mono .Wav file. 
The recording software displayed the text for them one sentence at a time and indicated the required emotion. They then recorded the sentence onto the phone. When they finished speaking, the recorded audio file was saved. They then recorded the same sentence with the same emotion a second time. The recording was done at the School of Information Science and Technology, Northwest University, in a quiet room in order to obtain speech signals with minimum noise. The distance between the speaker's mouth and the microphone of the mobile phone was 25 cm. We used the Audacity audio editing software \cite{audacity:xxx} to reduce the background noise of the speech signal. Audacity is very reputable and is also open source. 

\textbf{Data creation:}
We adopted a semi-supervised recording approach: The participants were given a short description of the purpose of the study and the recording system, and were then allowed to choose a convenient time to conduct the recording. 
Each participant was first asked to record all 25 emotion-specific sentences. They were provided with the sentence and the required emotion, for example, `Turn away and let me not see you again' to be spoken in an Angry way. Each recording was made twice. Next on the list could be the sentence `I think he was a thief who knocked on the door', to be spoken in a Fearful way, and so on. For these emotion-specific sentences, the required emotion was always the same for a given sentence, for all participants. After the 25 sentences, the participant was then asked to record the two common sentences, for example `my sister is coming on a plane'. Each common sentence was recorded twice for each of the five emotions, Neutral, Fearful, Happy, Sad and Angry. So they would first record the sentence in a Neutral way (twice), then record the same sentence in a Fearful way (twice) and so on. Participants always spoke in their own Amharic dialect. A complete set therefore comprised $25 \times 2 = 50$ recordings of emotion-specific sentences, and $2 \times 5 \times 2 = 20$ recordings of common sentences, 70 recordings in all.

\textbf{Judgments:}
Every recording was independently reviewed by all eight judges. Concerning emotion-specific sentences, each judge decided whether the recording expressed the emotion adequately or not, and made a binary decision, Accept or Reject. For the common sentences, the judge did not know the intended emotion of the recording. They decided which emotion the recording expressed. If it was unclear, their decision was Reject, otherwise their decision was one of the five emotions.

For emotion-specific sentences, a recording was only accepted for inclusion in the ASED dataset if five or more judges returned the decision Accept. Similarly, a common sentence was only accepted if five or more judges assigned the same emotion to it.

Because there were 65 partipants each of whom made 70 recordings, we would expect ASED to contain 4,550 recordings. However, because of the above selection process, many of these were rejected, resulting in 2,474 recordings in the dataset (Table \ref{no-utterances-per-speaker}).

\textbf{Inter-annotator agreement:} 
Since we had eight judges, the Fleiss kappa \cite{randolph2005free} coefficient was used to calculate the pairing agreement between participants.
\begin{equation} \label{eq:1}
\kappa=\frac{\overline{p}_{0}-\overline{p}_{e}}{1-\overline{p}_{e}} 
\end{equation}
The factor ${1-\overline{p}_{e}}$ gives the degree of agreement that is attainable above chance, and ${\overline{p}_{0}-\overline{p}_{e}}$ gives the degree of agreement actually achieved above chance: $k=1$, if all the raters are in complete agreement. Evaluation of the inter-rater agreement for our dataset in terms of Fleiss kappa is 0.8. This value shows a high agreement among our eight raters.

\textbf{Files and labelling:}
Each file was then labeled in the form a5-02-01-12.wav The first part of the name indicates the emotional state (n1 = neutral, f2 = fearful, h3 = happy, s4 = sad, a5 = angry), the second part indicates the sentence number (01-07), the third part indicates the repetition (01 = 1st repetition, 02 = 2nd repetition) and the fourth part indicates the anonymised participant ID (01 to 65).

The final dataset consists of 2,474 recordings, each between two and four seconds in length, 522 neutral, 510 fearful, 486 happy, 470 sad, and 486 angry. Recorded phrases were stored in five different folders, one for each emotion.
Table \ref{no-utterances-per-speaker} shows the final breakdown of utterances across emotions and dialects. The ASED dataset is evenly distributed across all emotion classes, as shown in Fig. \ref{ased-distribution}.  
ASED was split into training and testing sets randomly. The training set contains 90\% of the whole dataset. The test set contains the rest of the data.
The ASED dataset is freely available for research purposes\footnote{\url{https://github.com/Ethio2021/ASED_V1}}.

\subsection{Comparison} \label{comparison}
Before presenting our experiments, we briefly compare ASED to nine other datasets (Table \ref{ser-datasets}) in terms of parameters and creation methods. There are four for English, and one each for Chinese, German, Greek, Gujarati, and Hindi.

AESDD \cite{vryzas2018speech} is for Greek, uses five emotions and contains 19 emotionally neutral utterances derived from theatrical plays plus one improvised utterance. Each of these 20 is spoken in all five emotions by each actor. There are five professional actors and 500 utterances in total. No quality control or annotation process is mentioned. Recording is done at 44.1 kHz and 16 bits. Recordings were made at the sound studio of the Laboratory of Electronic Media.

CHEAVD \cite{li2017cheavd} is for Mandarin Chinese and uses six basic emotions plus an additional 20 emotions. There are 2,600 segments selected from 34 films, two tv series, two tv shows, one speech and one talk show. Each segment involves one speaker and there are 238 different speakers in all. Thus there are no recruited participants in this dataset, and all emotions are naturally occurring. Annotation of emotions is carried out by four judges. Pairwise kappa coefficients range between 0.41 and 0.58. However, there are 26 emotions rather than the usual six, so this accounts for the low agreement level. Recordings are all from pre-existing films etc and are at 44.1 kHz and 16 bits.

EGSC \cite{tank2020creation} is for Gujarati, uses six emotions and is based on 24 individual words. Each recording is for just one word, spoken with one of the emotions. There are nine speakers who are experts in drama, and about 1,296 utterances in total. Recording is done at 44.1 kHz using mobile phones. A quiet room was used.

EMO-DB \cite{burkhardt2005database} (as already discussed) is for German, uses five emotions and contains ten everyday sentences, five made of one phrase, five made of two phrases. There are ten speakers, nine qualified in acting and about 800 raw utterances in total. All recordings were judged by twenty listeners, who found that about 300 of the 800 utterances had a recognition rate greater than 80\%. 535 utterances were finally selected for the database. Recording is done at 16 kHz and 16 bits, and was carried out in the Anechoic chamber of the Technical Acoustics Department at the Technical University Berlin.

IEMOCAP \cite{busso2008iemocap} is for English, uses five emotions and contains three scripts selected from plays, plus improvised emotional dialogues. Seven speakers are professional actors, three are students. Recordings were judged by six evaluators using a majority voting system. There are 10,039 dialogue turns in the dataset (5,255 scripted turns and 4,784 spontaneous turns). Recording is done at 48 kHz and 16 bits, and took place in the Speech Analysis and Interpretation Laboratory (SAIL) at the University of Southern California (USC).

ITKGP-SEHSC \cite{koolagudi2011iitkgp} is for Hindi, uses eight emotions and contains 15 sentences. There are ten professional artists who each record every emotion on every sentence, all in one session. The total number of utterances in the database is 12,000. Recording was done at 16 kHz and 16 bits, working in a quiet room. Recordings were judged by 25 postgraduate and research students of IIT Kharagpur.

RAVDESS \cite{livingstone2018ryerson} (as already discussed) is for English, uses eight emotions and contains just two sentences. The 24 speakers are professional actors. Interestingly, emotion in this dataset is `self-induced' \cite{stanislavski1936actor}, rather than Acted. Moreover, there are two levels of each emotion. There are 4,320 utterances.  Project investigators first selected the best two clips for each speaker and each emotion. The selected recordings were then judged by 247 naive judges. The average Fleiss Kappa inter-rater score for speech was 0.57. Recording was at 48 kHz and 16 bits, and it was carried out in a professional recording studio at Ryerson University.

SAVEE \cite{jackson2014surrey} is for English, uses six emotions and contains three common sentences plus two emotion-specific sentences and ten generic sentences (different for each emotion and phonetically-balanced).  There are four speakers (postgraduates and researchers, not professional actors) and there are 480 utterances in total. Recordings were judged by ten evaluators (all students) and average classification accuracy was 66.5\%. Recording was at 44.1 kHz and 16 bits, and took place in the 3D vision laboratory at University of Surrey.

TESS \cite{SP2_E8H2MF_2010} \cite{dupuis2011recognition} is for English, uses seven emotions and contains 200 individual words. These are spoken in a carrier sentence `Say the word X'. There are two female speakers who are professional actors, and 2,800 utterances. Fifty-six judges identified the emotion in each recording with an average accuracy of 82\% \cite{dupuis2011recognition}. `Pleasant' was the least well recognised, and `Anger' the most. Recording was at 24.4 kHz and 16 bits, and took place in a sound attenuating booth at University of Toronto.

A number of interesting comparisons can be made between ASED and the datasets mentioned above. Firstly, concerning participants, we can see that mostly professional actors are used for the recordings. The only exceptions are SAVEE which uses members of the campus community, and CHEAVD which has no participants as such, but uses pre-occurring speech in tv shows etc. For ASED we have a mixture of postgraduates, undergraduates and business persons making the recordings. It is an interesting assumption among all the papers that actors are the best choice. However, the way in which emotions are portrayed in drama is surely entirely different from everyday life. Actors exaggerate and distort the facets of the emotion they portray. Usually they also do this in a way which has come to be expected within the particular entertainment genre, but which is not real. So we would argue that alternatives should be explored if the ultimate goal is to train SER systems for practical tasks such as detecting an upset customer in a call center. The number of participants used for the datasets varies dramatically from just two (TESS) up to 231 (CHEAVD). Excluding CHEAVD, the average is 9.2, so the use of 65 speakers in ASED is well above average, covers all the main dialects in Ethiopia and can thus be considered a representative sample.

For emotions, mostly the standard five are chosen as the basis and ASED conforms to this norm. Expression of the emotions is generally of the Acted form, i.e. simply asking the participant to speak in, say, an angry way, and leaving it to them how to do it. However, RAVDESS uses self-inducement [31] whereby the actor conjurs up the emotion in themselves over a period of some minutes before speaking. IEMOCAP uses scripts selected from plays, where the emotion is developed in the dialogue over several turns. They also try improvised dialogues based around an idea, e.g. for `sad' the idea is that a close friend has recently died and one person is conforting another. This can allow the emotion to develop more naturally. At the other extreme, TESS expresses the emotion in a single word, with no preparation. For ASED, we use the simulated approach (e.g. please speak in a sad way) but we rely on our quality control mechanism to ensure that the results are satisfactory.

For the choice of sentences which participants must speak, this varies greatly. EGSC uses single words, while TESS uses `Say the word X' where X is a single word. RAVDESS has just two neutral sentences, spoken with different emotions. EMO-DB has ten, ITKGP-SEHSC 15, AESDD 20 and SAVEE 120; by contrast, IEMOCAP is using drama scripts plus dialogues, so this is a more naturalistic approach. At the extreme, CHEAVD has 2,600 tv show scripts, so this is an extrapolation of the IEMOCAP idea. For ASED, we are using 25 emotion-specific sentences plus two neutral sentences which are common to all emotions, so this is closest to EMO-DB and AESDD.

Considering the amount of data produced, ASED (2,474 utterances) compares well with the other datasets; the minimum is SAVEE (480) the maximum ITKGP-SEHSC (12,000 dialogue turns) and the average is 3680. So, for a first Amharic dataset, ASED provides enough data to perform basic training tasks, as we show later.  Finally, we consider the judges and agreement level. The number of judges used for CHEAVD (4), IEMOCAP (6) and SAVEEE (10) are comparable to ASED (8). EMO-DB (20), ITKGP-SEHSC (25), TESS (56) and RAVDESS (247) use greater numbers. Considering the sampling frequency, both EMO-DB and ITKGP-SEHSC use 16 kHz which is the same as ASED. The majority of datasets are recorded in a quiet laboratory environment, the approach adopted for ASED as well.

The eight ASED judges showed a high level of inter-rater agreement (Fleiss kappa 0.8 -- see Section \ref{creation-of-ased} above). RAVDESS report a kappa of 0.57 over their 247 judges, while CHEAVD report kappa 0.41-0.58 over four judges and 26 emotions. Obviously, these figures are not directly comparable, for example the CHEAVD task is much harder because there are many more emotions, however, the ASED agreement level seems very good.

In conclusion, the proposed ASED SER dataset for Amharic appears to compare well with the others we mention here. In the following sections, we present some initial experiments where the ASED data is used for Amharic SER.
	
\section{FEATURE EXTRACTION FOR SER}
The speech signal contains a large number of parameters reﬂecting emotional characteristics. One of the key issues within SER research is the choice of features which should be used.
  After reviewing many works on emotion recognition, it is clear that Mel-spectrograms and Mel-Frequency Cepstral Coefficients (MFCC) are broadly utilized in audio classification and speech emotion recognition \cite{issa2020speech}. We briefly review these methods below.

\subsection{Mel-Spectrograms}
\begin{figure}[htbp]
			\begin{subfigure}{.19\textwidth}
				\includegraphics[width=.8\linewidth]{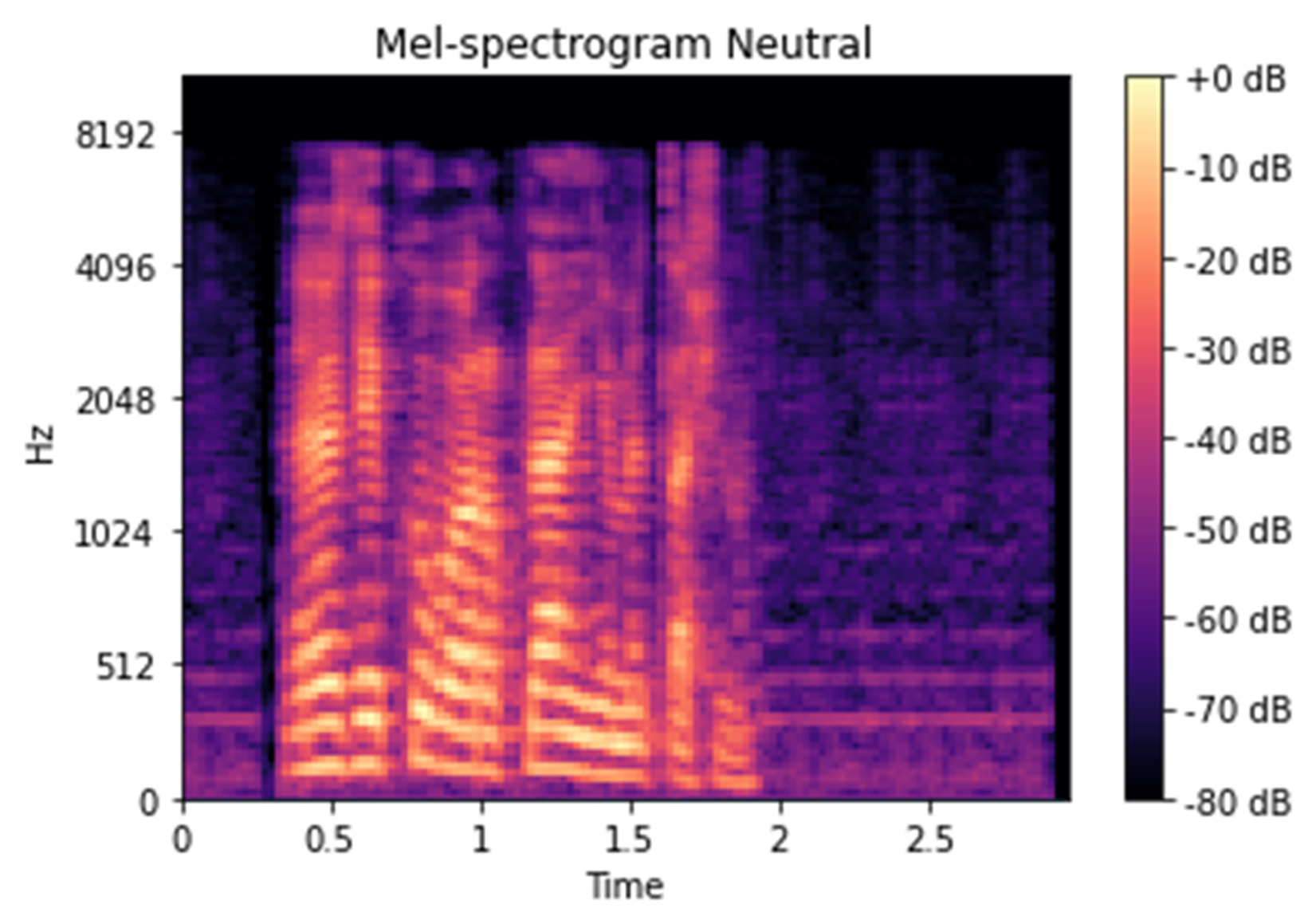}
				\caption{Neutral}
			\end{subfigure}
			\begin{subfigure}{.19\textwidth}
				\includegraphics[width=.8\linewidth]{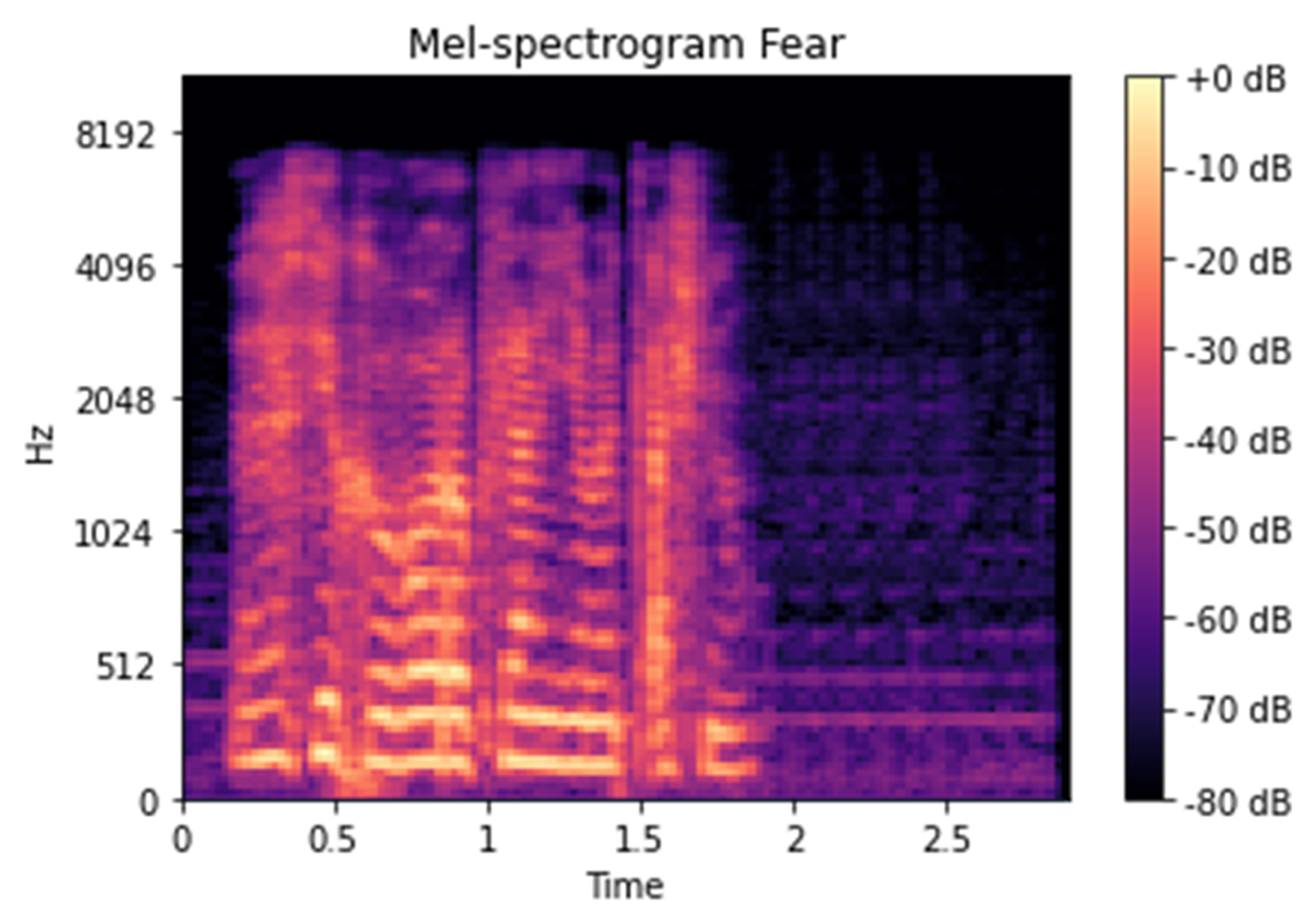}
				\caption{Fearful}
			\end{subfigure}
			\begin{subfigure}{.19\textwidth}
				\includegraphics[width=.8\linewidth]{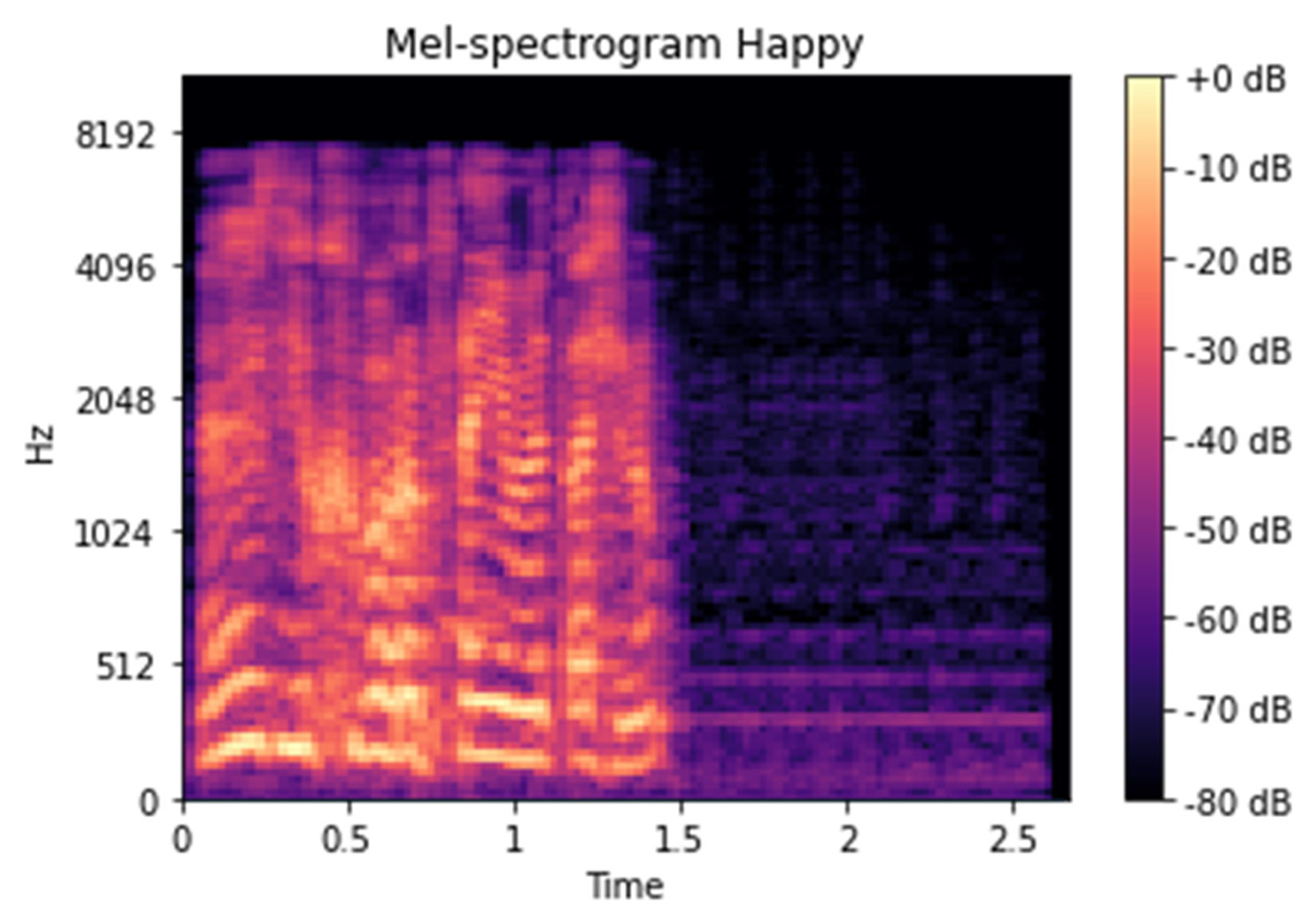}  
				\caption{Happy}
			\end{subfigure}
			\begin{subfigure}{.19\textwidth}
				\includegraphics[width=.8\linewidth]{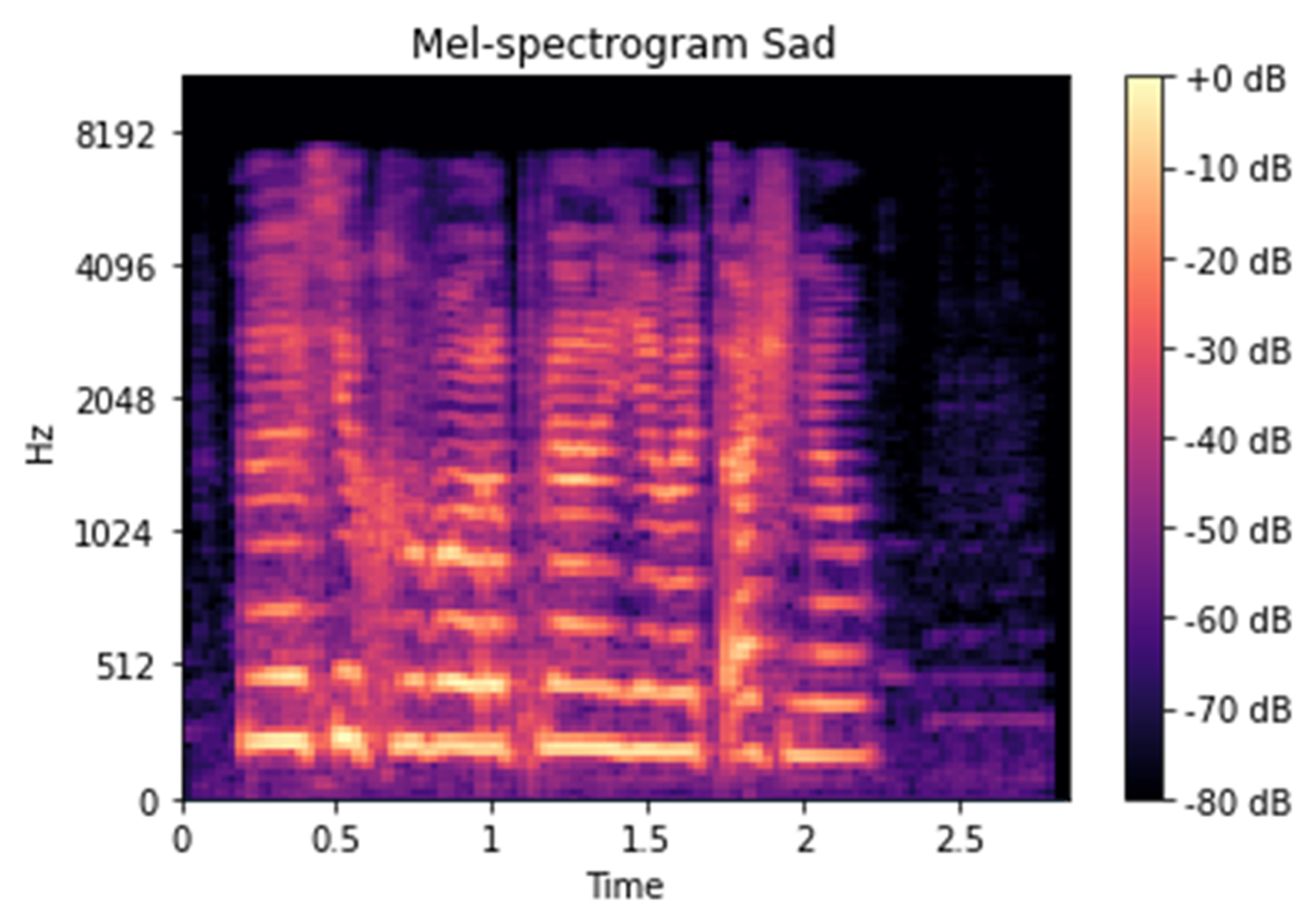}
				\caption{Sad}
			\end{subfigure}
			\begin{subfigure}{.19\textwidth}
				\includegraphics[width=.8\linewidth]{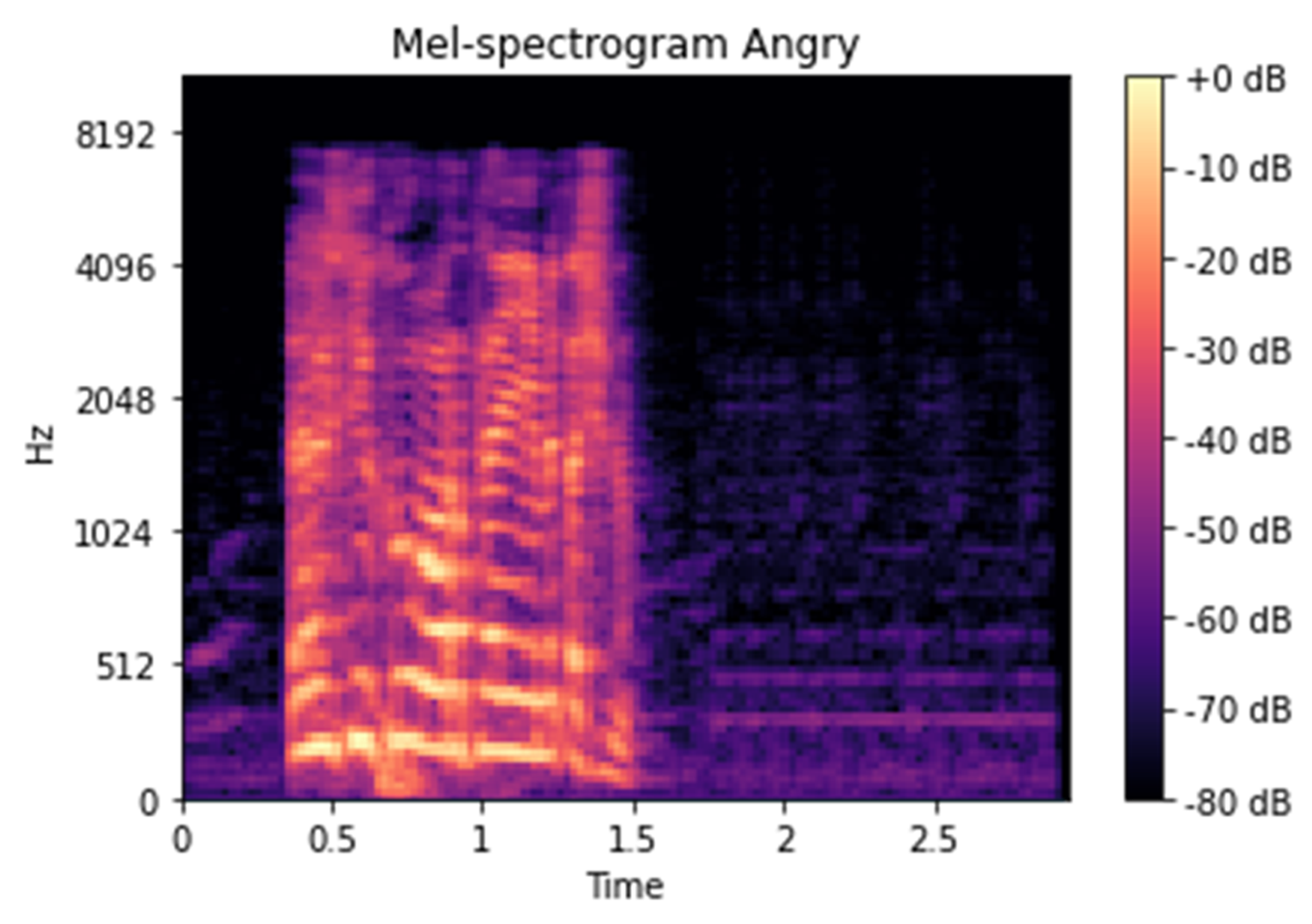}  
				\caption{Angry}
			\end{subfigure}
			\caption{Examples of Mel-spectrogram features for each emotion in ASED.}
			\label{mel-plots}
		\end{figure}

The spectrogram is the relationship between the time and frequency of the audio signal. Different emotions can show different patterns in the energy spectrum. The Mel-spectrogram represents the audio signal in Mel-scale. The logarithmic form of the Mel-spectrogram can be better understood because humans perceive sound on a logarithmic scale. 
The human ear is observed to act as a sub-band filter bank.
These filters overlap and are unevenly spaced on the frequency axis. In audio processing, the signal is considered stationary within 10 to 30 ms, and therefore a window with a shorter duration is selected \cite{tak2017novel}.
Sample Mel-spectrogram plots for each of the five emotions are shown in Fig. \ref{mel-plots}.
\subsection {Mel-Frequency Cepstral Coefficients (MFCC)} 
\begin{figure}[htbp]
			\begin{subfigure}{.19\textwidth}
				\centering
				\includegraphics[width=.8\linewidth]{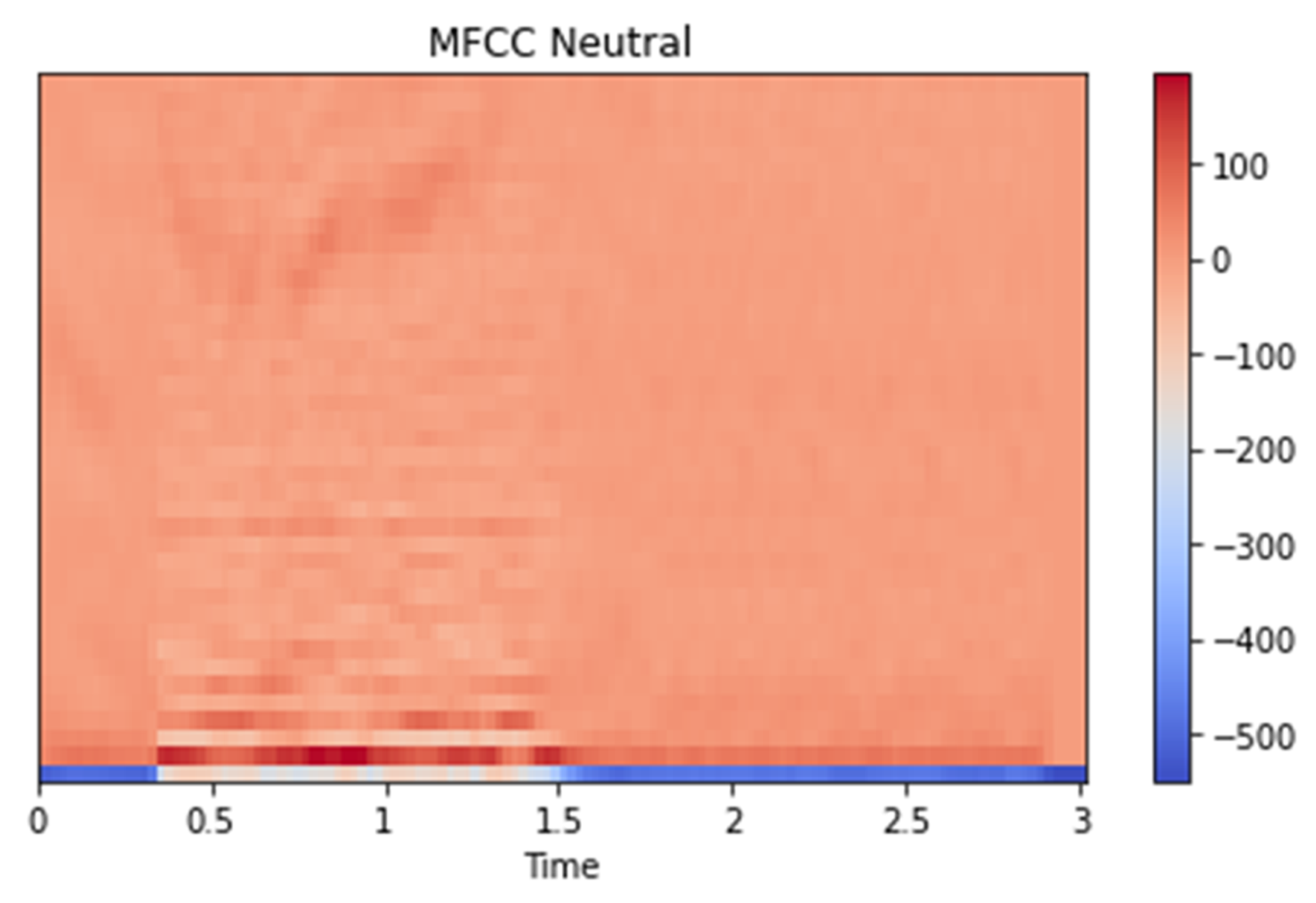} 
				\caption{Neutral}
			\end{subfigure}
			\begin{subfigure}{.19\textwidth}
				\centering
				\includegraphics[width=.8\linewidth]{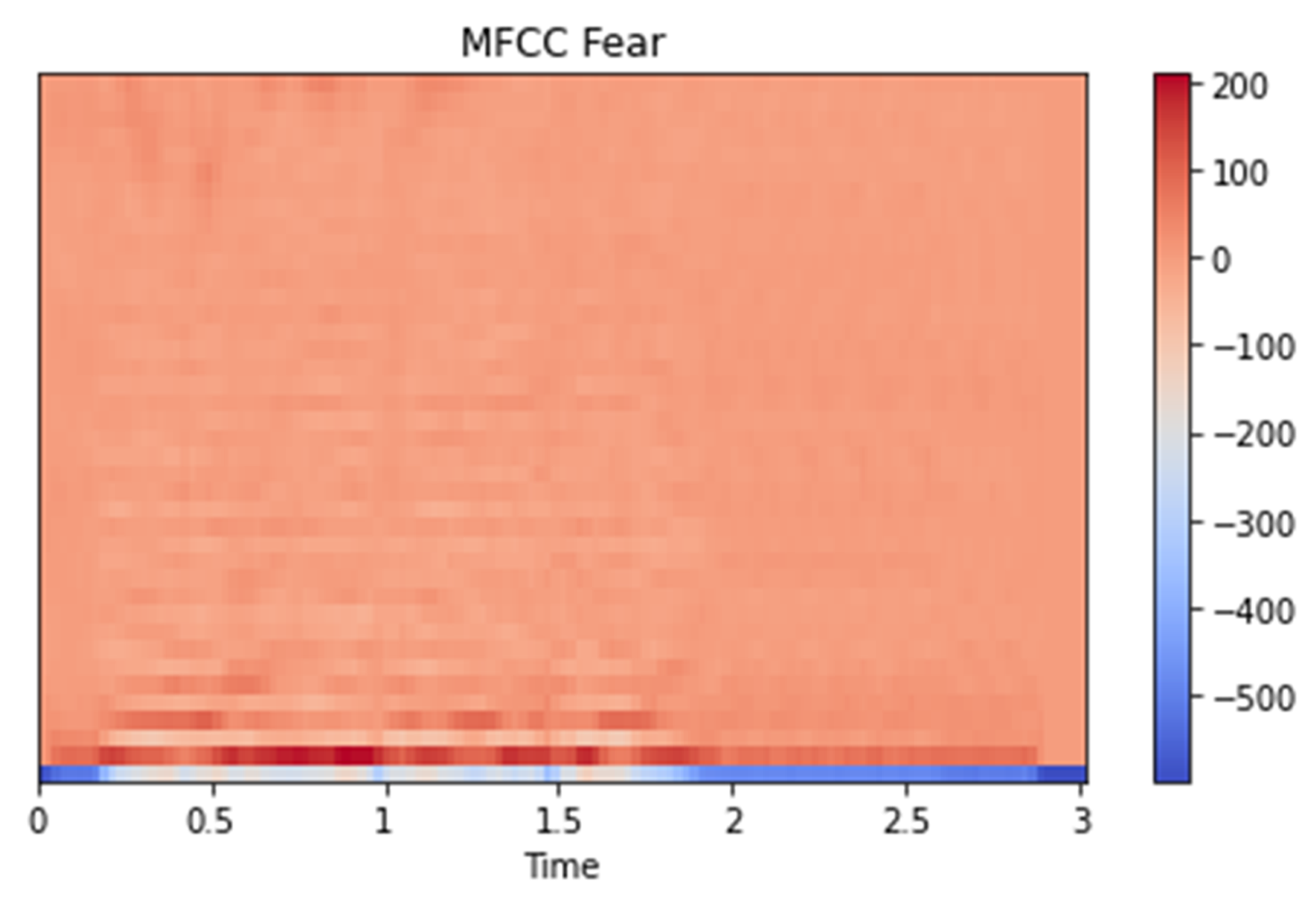}
				\caption{Fearful}
			\end{subfigure}
			\begin{subfigure}{.19\textwidth}
				\centering
				\includegraphics[width=.8\linewidth]{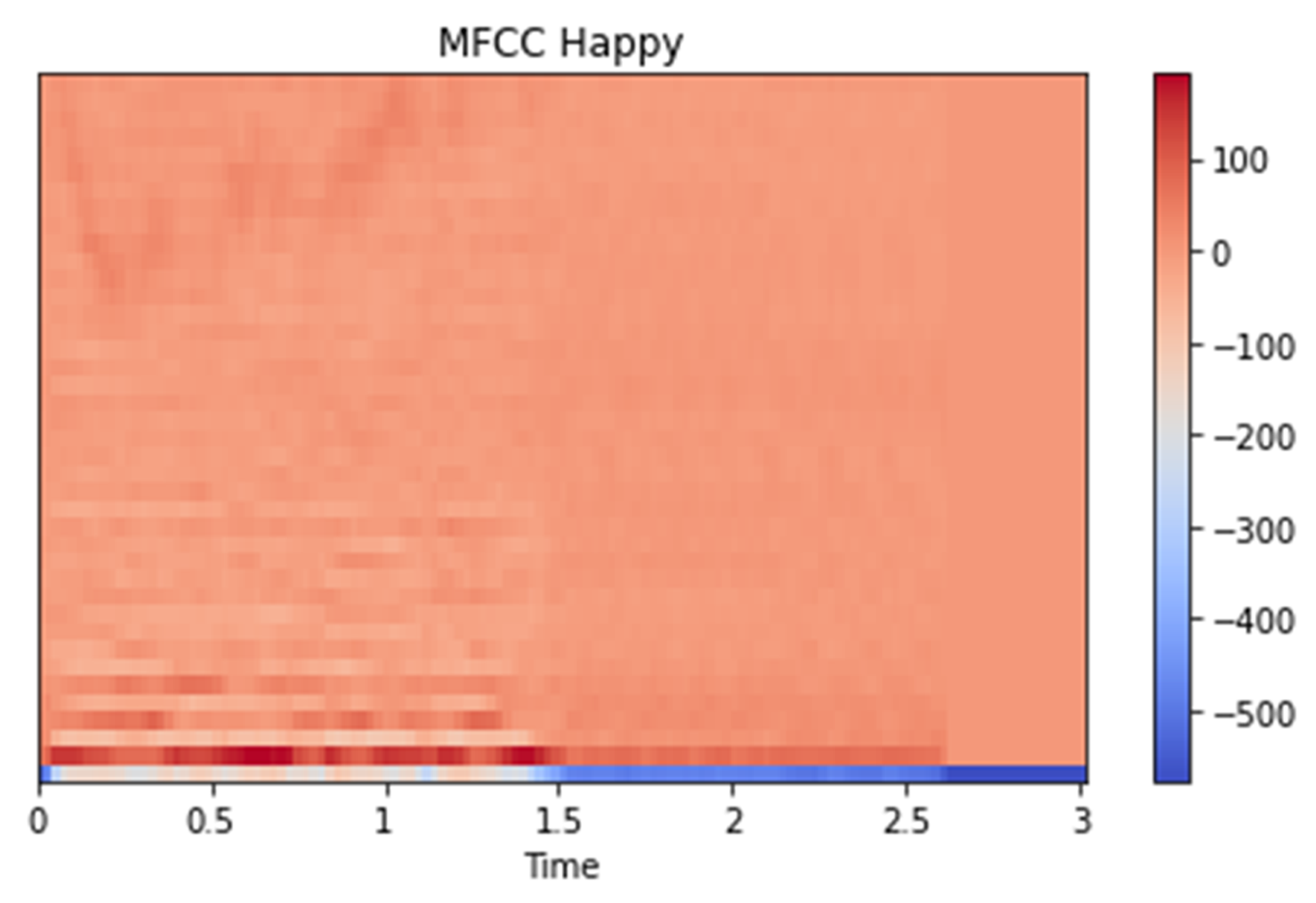} 
				\caption{Happy}
			\end{subfigure}
			\begin{subfigure}{.19\textwidth}
				\centering
				\includegraphics[width=.8\linewidth]{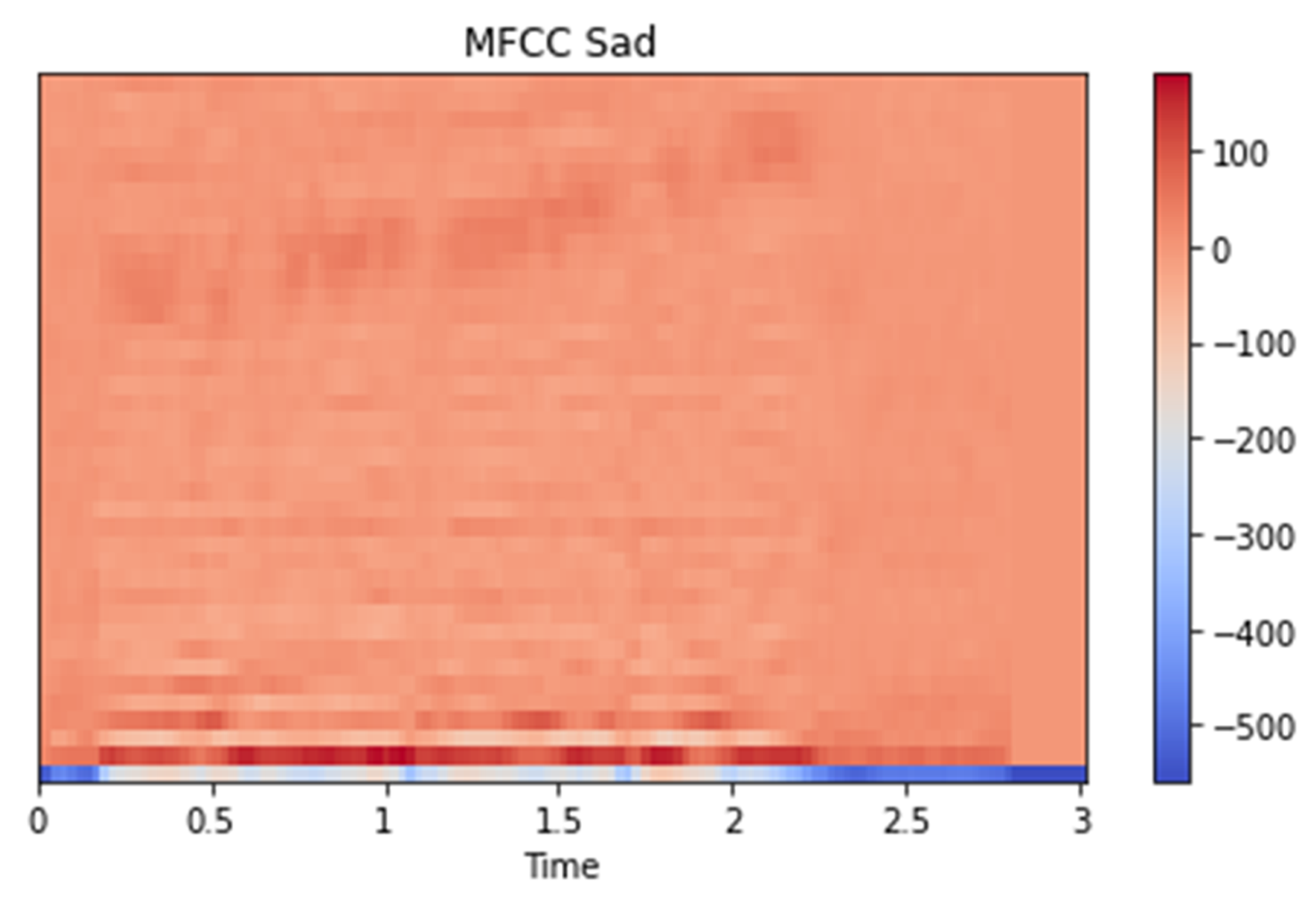}  
				\caption{Sad}
			\end{subfigure}
			\begin{subfigure}{.19\textwidth}
				\centering
				\includegraphics[width=.8\linewidth]{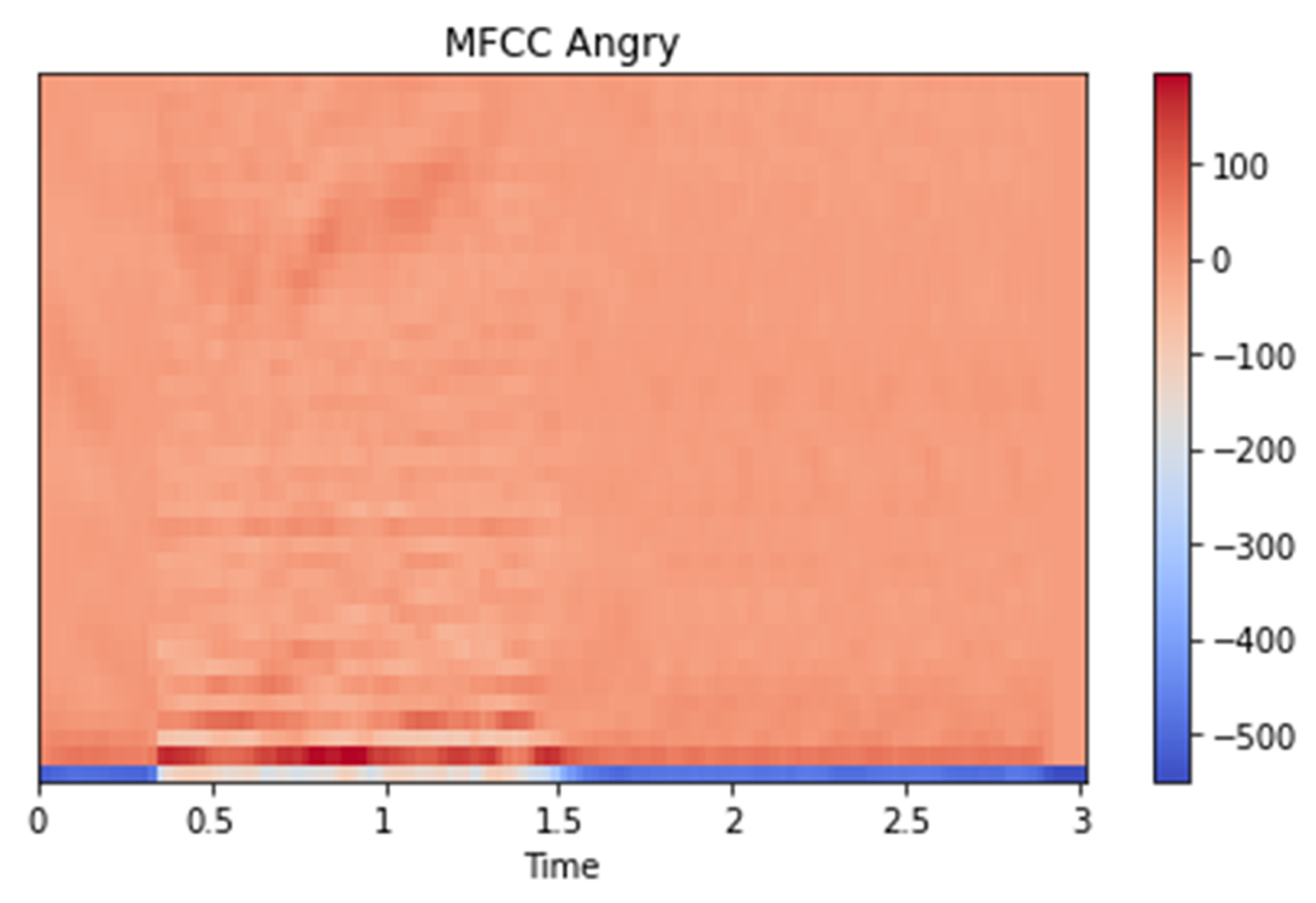}  
				\caption{Angry}
			\end{subfigure}
			\caption{Examples of MFCC features for each emotion in ASED.}
			\label{mfcc-plots}
		\end{figure}

        Mel-Frequency Cepstral Coefficient (MFCC) is a coefficient that expresses the short-term power spectrum of a sound. It uses a series of steps to imitate the human cochlea, thereby converting audio signals. The Mel scale is significant because it approximates the human perception of sound instead of being a linear scale \cite{shaw2016emotion}. Sample MFCC plots for each of the five emotions are shown in Fig. \ref{mfcc-plots}.
     
\section{NETWORK ARCHITECTURES AND SETTINGS} \label{architectures}
\subsection{Deep learning architectures}
As discussed earlier, most of the previous studies employ CNN-based models for SER \cite{kwon2020cnn}. Among such models, the notable ones include Alex-Net  \cite{krizhevsky2012imagenet,sajjad2020clustering}, VGG \cite{simonyan2014very,molchanov2016pruning}, and ResNet50 \cite{he2016deep,george2017deep}, as well as LSTM \cite{kumbhar2019speech}.
\begin{table}[ht]
	    \centering
        \caption{VGGb architecture used in experiments.}
			\begin{tabular}{c}
				\\\hline  
				Input 40 x 174 x 1\\ \hline
				Conv2D 1 (32 filters + kernel size = 2 + Relu) \\
				MaxPooling 2D (2,2) + Dropout (0.15)  \\\hline 
				Conv2D 2 (64 filters + kernel size = 2 + Relu)\\ 
				MaxPooling 2D (2,2) + Dropout (0.15) \\\hline 
				Conv2D 3 (128 filters + kernel size = 2 + Relu)\\ 
				MaxPooling 2D (2,2) + Dropout (0.15) \\\hline 
				Conv2D 4 (256 filters + kernel size = 2 + Relu)\\ 
				MaxPooling 2D (2,2) + Dropout (0.15) \\\hline 
				AveragePooling2D Global\\ 
				Dense (64, Relu) \\\hline 
				Dense (Soft Max 5) 
				\\\hline 
			\end{tabular}
			\label{vgg-style-network}
\end{table}
VGG is one of the earliest CNN models used for signal processing. It is well known that the early CNN layers capture the general features of sounds such as wavelength, amplitude, etc., and later layers capture more specific features such as the spectrum and the cepstral coefficients of waves. This makes a VGG-style model suitable for the SER task. After some experimentation, we found that a model based on VGG but using four layers gave the best performance. We call this proposed model VGGb and used it for our experiments. Table \ref{vgg-style-network} shows the settings for VGGb. Alex-Net, ResNet50 and LSTM were also used for comparison.
\subsection{Experimental setup}
The standard code for the Alex-Net, VGG, ResNet50 and LSTM models was used for the experiments. The VGG code was adapted to create VGGb (Table \ref{vgg-style-network}). For the other models, the original network configuration and parameters were used.

We used the Keras deep learning library, version 2.0, with Tensorflow 1.6.0 backend \cite{kumbhar2019speech} to build the emotion recognition models. The models were trained using a 2.30 GHz (CPU) Intel(R) Xeon(R) CPU. The Adam optimization algorithm was used to train our model, with categorical cross-entropy as the loss function; training stopped after 100 epochs, and the batch size was set to 16.
\section{Experiment 1: Choice of Features for Amharic SER}
\subsection{Outline}
As we have mentioned, Mel-spectrograms and MFCC are two forms of feature which are widely used within SER systems for other languages. We therefore wished to determine which of these was most suitable for Amharic SER. The experiment was divided into four parts. First, a direct comparison of Mel-Spectrograms and MFCC was carried out, by substituting each one in turn into the VGGb architecture and determining the resulting SER performance. Second, the dataset sentences used for training and testing were varied in order to show that this did not affect results. Third, the dialects used for training and testing were varied. Fourth, the groups of speakers used for training and testing were varied. The effect of steps 2-4 was to validate the performance of the two feature types by performing a kind of cross-validation based on sentences, dialects and speaker groups. We now describe each experiment in turn.

\subsection{Experiment 1.1: Initial Comparison}
We adopted the VGGb model for our experiments. The main settings are shown in Table \ref{vgg-style-network}. Training and testing were performed with ASED data. We extracted both Mel-spectrogram and MFCC features, using the librosa v0.7.2 library \cite{sharmin2020bengali}. Mel-spectrogram was extracted with 128 bands, and MFCC with 40 bands, according to the standard settings of the tool.

First, the model was trained and evaluated using just Mel features. Second, it was trained and evaluated using just MFCC features. Data was split 90\% train and 10\% test. The model was trained five times with a different random train/test split, and the average result was reported.

The results are shown in Table \ref{recog-accuracy}. MFCC has proven to be effective at extracting the important features \cite{livingstone2018ryerson,sajjad2020clustering,molchanov2016pruning}, as has been demonstrated in previous experiments with VGG CNN-based SER models. Our results with VGGb confirmed this trend. The classification accuracy was 90.73\% for MFCC as compared with 81.05\% for Mel-spectrograms.

\begin{figure}[ht]
			\begin{subfigure}{.4\textwidth}
				\centering
				\includegraphics[width=.6\linewidth]{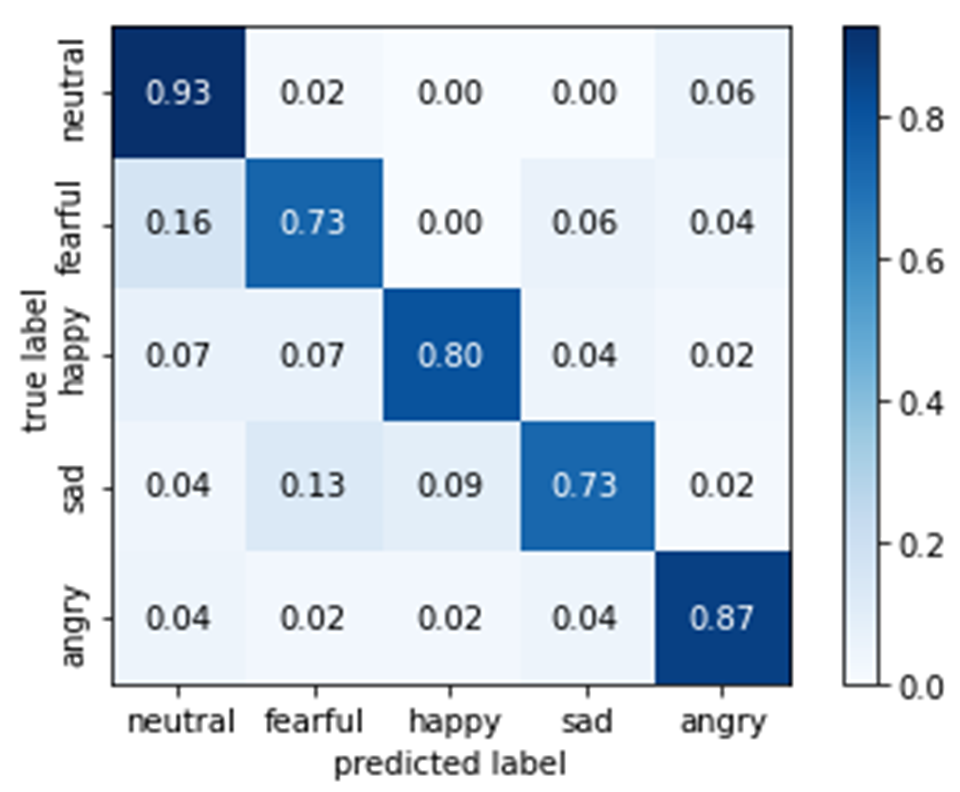}  
				\caption{Mel-Spectrogram}
			\end{subfigure}
			\begin{subfigure}{.4\textwidth}
				\centering
				\includegraphics[width=.6\linewidth]{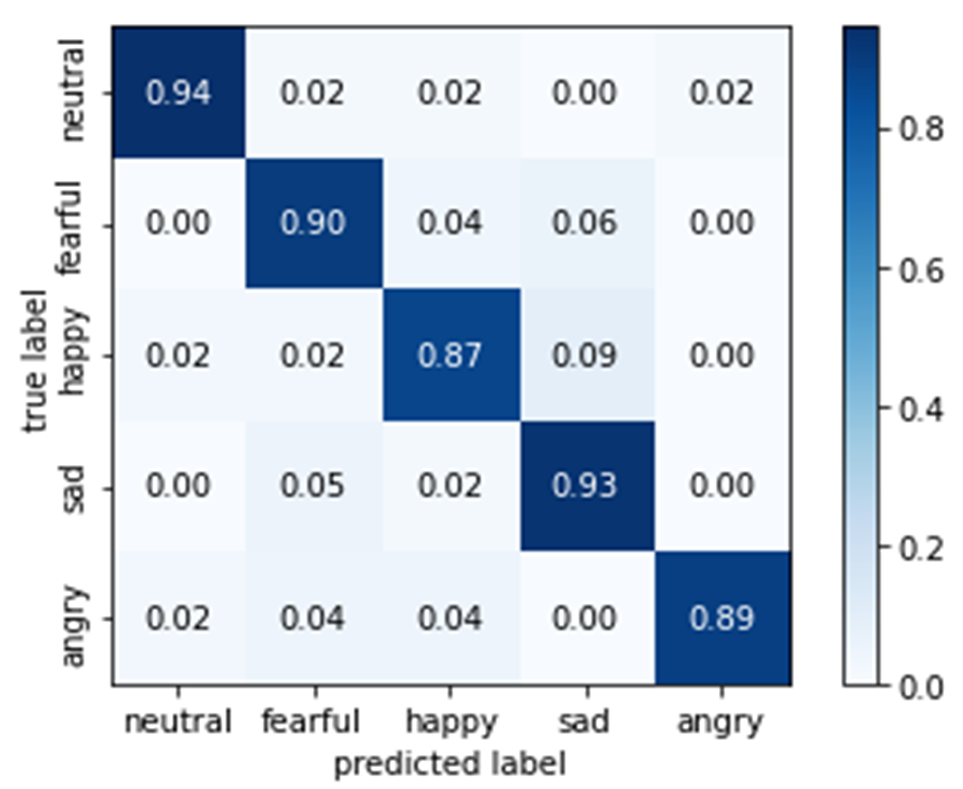}  
				\caption{MFCC}
			\end{subfigure}
			\caption{VGGb confusion matrices using Mel-Spectogram and MFCC features on  text-dependent ASED.}
			\label{confusion-matrices}
	 	\end{figure}

The confusion matrix for each feature is shown in Fig. \ref{confusion-matrices}. As shown in Fig. \ref{confusion-matrices}(a), the Mel-spectrogram VGGb model incorrectly classifies 13\% of sad cases as fearful. In consequence, prediction for the sad class is reduced to 73.0\%. The VGGb model over Mel-spectrogram, Fig. \ref{confusion-matrices}(a), shows less prediction gains in predicting the fearful as neutral (16.0\%) when compared to MFCC (73.0\% to 90.0\%).
This outcome appears to be conceivable because the MFCC can benefit from the diﬀerence between the word distributions of fearful and neutral expressions. It is striking that the VGGb model using Mel-spectrogram incorrectly predicts the happy class as fearful and neutral 7.0\% of the time, sad as fearful 13.0\% of the time, and sad as happy 9.0\% of the time, even though these emotional states are opposite to each other.
        
Compared to the Mel-spectrogram, the MFCC-based model in Figure \ref{confusion-matrices}(b) shows significant gains when predicting the sad class (93\% vs. 73\% for Mel) and the fearful class (90\% vs. 73\% for Mel). Here, MFCC incorrectly predicts instances of the sad class as happy only 2.0\% of the time, compared to 9\% for Mel-spectrogram. The frequency of incorrect cases fearful-to-neutral and sad-to-fearful relative to the Mel-spectrogram decreased from 16.0\% to 0.0\%, and from 13.0\% to 5.0\% respectively. The values on the diagonal axis indicate that the accuracy of the correctly predicted class has increased.

The Confusion matrices show that certain emotions are often mistaken for other emotions. Likewise, a few emotions are easier to recognize. This could be because the test data is based on participants pretending to speak with designated emotions, and people found it difficult to imitate certain emotions. It can be seen that neutral, fearful, sad, and angry are the emotions that are less difficult to recognize in Amharic speech as opposed to happy which is the most difficult.

Experiment 1.1 suggested that MFCC features were superior to Mel-spectrogram, but it did not indicate why this was the case, or what experimental factors could influence the outcome. Therefore, three further variations of the experiment were performed to investigate this further.
 \FloatBarrier 
\begin{table}[ht]
        \centering
        \caption{Accuracy of VGGb on ASED using  MFCC and Mel-Spectrogram Features.}
        \begin{tabular}{lccl}
        \hline
        Dataset & \multicolumn{2}{c}{Features} & Approach \\\hline
        \multirow{2}{4em}{ASED} & Mel-Spectrogram    & MFCC    &  \multirow{2}{4em}{VGGb} \\\cline{2-3}
        & 81.05 & 90.73 & \\ \hline         
        \end{tabular}
        \label{MFCC-Mel}
        \end{table}
         \FloatBarrier 
         
          \FloatBarrier 
 \FloatBarrier 

\subsection{Experiment 1.2: Independence of Sentences}
Recall that the dataset consists of five sentences, one for each emotion, and two common sentences, which are spoken in every emotion. In this experiment sentences were either used for training or for testing according to the scheme in Table \ref{recog-accuracy}. This was to determine how dependent the recognition of emotion is on the sentence used to express it. For each of the schemes shown in the table, VGGb was trained once using Mel-spectrograms and once using MFCC. The average result for each feature type is shown at the bottom.
As can be seen in Table \ref{recog-accuracy}, the average performance for MFCC (80.37\%) is higher than that for Mel-spectrogram (72.12\%). The standard deviations for these values are 1.60 and 3.42 respectively. In each training scenario, MFCC is better than Mel-spectrogram; moreover, the standard deviations are low, indicating that there is only a small difference in performance between different scenarios, i.e. choices of sentences for training and testing.

 \FloatBarrier 
 \begin{table}[ht]
        \centering
        \caption{Accuracy of VGGb on ASED using  MFCC and Mel-Spectrogram Features: Sentence Independence.}
           	 \begin{tabular}{cccccc}
           		 \hline  
           	Training Sentences & Testing Sentences & Mel-Spectrogram & MFCC  \\\hline 
           	1+2+3+4+5  	    &  6+7  &  75.51\%  &82.10\%\\
            1+2+3+6+7  	& 5+4  &   75.54\%   &81.08\% \\
            1+4+5+6+7  	& 2+3  &   69.46\%   &78.37\% \\
            2+3+4+5+6  	& 1+7 &    72.73\%   &79.91\% \\\hline
            \textbf{Average} && \textbf{73.31\%} & \textbf{80.37\%} \\
            \textbf{St. Dev.} && \textbf{2.89} & \textbf{1.60} \\
           		\hline 
           	\end{tabular}
           	\label{recog-accuracy}
           \end{table} 
       \FloatBarrier

\subsection{Experiment 1.3: Independence of Dialects}
ASED consists of recordings in four dialects, Gojjam, Wollo, Shewa and Gonder. In this experiment, the model is trained with utterances spoken in three dialects, and then tested with utterances in the fourth dialect, according to the scheme in Table \ref{Dialects Independent}. Similar to Experiment 1.2, this was to determine how dependent the emotion found is on the dialect used to express it. In Table \ref{Dialects Independent}, the average performance for MFCC (74.51\%) is higher than that for Mel-spectrogram (67.29\%). The standard deviations for these values are 1.93 and 1.99 respectively, suggesting that dialect is only making small changes to the performance figures and is not affecting the overall superiority of MFCC under all training scenarios.

 \FloatBarrier 
    \begin{table}[ht]
        \centering
        \caption{Accuracy of VGGb on ASED using  MFCC and Mel-Spectrogram Features: Dialect Independence.}
           	 \begin{tabular}{cccc}
           		 \hline  
           	Training Dialects & Testing Dialect &  Mel-Spectrogram & MFCC  \\\hline 
           	Wollo+Shewa+Gonder  	&  Gojjam &  64.67\% & 72.03\%\\
            Gojjam+Shewa+Gonder 	& Wollo & 68.36\%  &75.32\% \\
            Wollo+Gojjam+Gonder 	& Shewa  & 69.22\%  &76.56\% \\
           	Wollo+Shewa+Gojjam  	& Gonder & 66.91\%  &74.12\%
            \\\hline
            \textbf{Average} && \textbf{67.29\%} & \textbf{74.51\%} \\
            \textbf{St. Dev.} && \textbf{1.99} & \textbf{1.93} \\
           		\hline 
           	\end{tabular}
           	\label{Dialects Independent}
           \end{table}    
  \FloatBarrier

\subsection{Experiment 1.4: Independence of Speaker Category}
ASED was recorded with three groups of speakers, Professional, Semi-professional and Amateur. To investigate how results might be affected by the speaker category, training was done with speakers from two groups and then tested with speakers of the third group, according to the scheme in Table \ref{Speaker-independent}. The results in that table once again show that performance for MFCC (76.91\%) is higher than that for Mel-spectrogram (71.57\%). The standard deviations for values are 1.53 and 2.50 respectively, indicating that the choice of speaker groups is not changing the result very much. Moreover, it is interesting that Amateur speakers do not give the lowest emotion recognition results. It is often stated that theatre actors should be used for generating emotion datasets, but the low standard deviations shown here do not lend support to that viewpoint.

Overall, Experiments 1.2-1.4 suggest that the advantage of MFCC over Mel-spectrogram as measured by emotion recognition accuracy is robust with respect to sentence choice, dialect and speaker category.

  \FloatBarrier 
   \begin{table}[ht]
        \centering
        \caption{Accuracy of VGGb on ASED using  MFCC and Mel-Spectrogram Features: Speaker Group Independence.}
           	 \begin{tabular}{cccc}
           		 \hline  
Training Groups & Testing Group & Mel-Spectrogram & MFCC  \\\hline 
Semi-professional + Amateur & Professional            & 74.36\% & 78.18\%\\
Professional + Amateur & Semi-professional            & 69.52\%  &75.21\% \\
Professional + Semi-professional & Amateur            & 70.84\%  &77.34\%
\\\hline
\textbf{Average} && \textbf{71.57\%} & \textbf{76.91\%} \\
\textbf{St. Dev.} && \textbf{2.50} & \textbf{1.53} \\
           		\hline 
           	\end{tabular}
           	\label{Speaker-independent}
           \end{table}  
 \FloatBarrier 
\section{EXPERIMENT 2: COMPARISON OF SER METHODS FOR AMHARIC}
Experiment 1 demonstrated, under four different measures, that MFCC features are better than Mel-spectrogram for use in Amharic SER. The aim of Experiment 2 was to determine which of the models discussed in Section \ref{architectures} would give the best performance when applied to ASED data using MFCC features.

Recall that the four models are: Alex-Net, VGGb, ResNet50 and LSTM. The network configuration for VGGb was the same as in the previous Experiment (Table \ref{vgg-style-network}). For the other models, the standard configuration and settings were used.
Once again, the librosa v0.7.2 library \cite{sharmin2020bengali} was used to extract MFCC features to input to the models. Each model was trained five times using a 90\%/10\% test/train split and the average results were computed.

  \FloatBarrier         
  \begin{table}[ht]
        \centering
        \caption{Comparison of VGGb with other CNN and RNN models, all applied to the ASED dataset.}
           	 \begin{tabular}{cccc}
           		 \hline  
           	No.	& Model& Training Time & Accuracy  \\\hline 
           	1	&  LSTM  &  00:33:45 &66.94\%\\
            2	& Alex-Net  & 10:21:08  &81.82\% \\
            3	&  ResNet50  & 08:44:15  &91.13\% \\
           	4	&  VGGb  & 00:31:18  &90.73\% \\
           		\hline 
           	\end{tabular}
           	\label{comparison-vgg-cnn-rnn}
           \end{table}    
           \FloatBarrier
\FloatBarrier
        \begin{figure}[ht]
			\centering
			\includegraphics[width=60mm,scale=1.0]{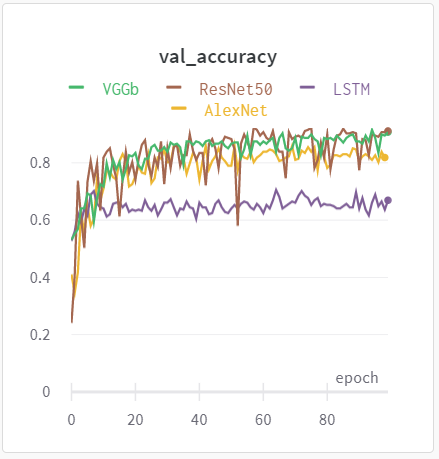}
			\caption{ Val-accuracy training curves for Alex-Net, ResNet50, LSTM and VGGb models on ASED.}
			\label{training-curves}
		\end{figure}
		\FloatBarrier
Results are presented in Table \ref{comparison-vgg-cnn-rnn}. Although ResNet50 had the highest accuracy (91.13\%), VGGb was just 0.4\% behind (90.73\%). Moreover, VGGb was much faster than ResNet50 (00:31:18 vs. 08:44:15), showing that it is much more efficient and hence more suitable for applying to large datasets.

The computational simplicity of VGGb in terms of training time and overall accuracy is again shown in Fig. \ref{training-curves} which represents the val-accuracy curve for the implemented models, AlexNet, ResNet50, LSTM and VGGb. The models are trained for 100 epochs. The curves show
that after the 20th epoch, the val-accuracy starts stabilizing. The curve for VGGb looks like a better fit, while that for ResNet50 shows more noisy movements than the other models.

\section{Experiment 3: COMPARISON OF ENGLISH, GERMAN AND AMHARIC SER}
The aim of Experiment 3 was to compare the performance of VGGb on three datasets, RAVDESS (English), EMO-DB (German) and ASED (Amharic).
Naturally there are interesting SER databases for many different languages (Table \ref{ser-datasets}). However,
in previous work, RAVDESS and EMO-DB have been extensively used for SER research, and are frequently quoted as baselines. This is why we chose to use these datasets, rather than others, for the comparison. Table \ref{comparison-ASED-RAVDESS-EMO-DB} (Section \ref{existing_ser_datasets}) provides summary information for the three datasets, which we also analysed in detail in Section \ref{comparison}. 

The VGGb model was trained using each of the three datasets. Network configuration and settings were there same as for previous experiments (Table \ref{vgg-style-network}). Input features were MFCC, extracted by librosa v0.7.2. In each case, training was carried out five times with a 90\%/10\% test train split and the average results were computed.
\FloatBarrier
        \begin{table}[h]
        \centering
        \caption{Comparison of VGGb applied to different datasets.}
           	 \begin{tabular}{cccc}
           		 \hline  
           	Model	& Dataset  & Training Time in min & Accuracy  \\\hline 
                	&  RAVDESS &00:12:33  &86.15\%\\
            VGGb 	& EMO-DB &00:06:02 &88.71\% \\
                	& ASED  &00:31:18  &90.73\% \\
           		\hline 
           	\end{tabular}
           	\label{vgg-accuracy-different-datasets}
           \end{table} 
           \FloatBarrier
\FloatBarrier
 \begin{figure}[ht]
			\centering
			\includegraphics[width=65mm,scale=1.5]{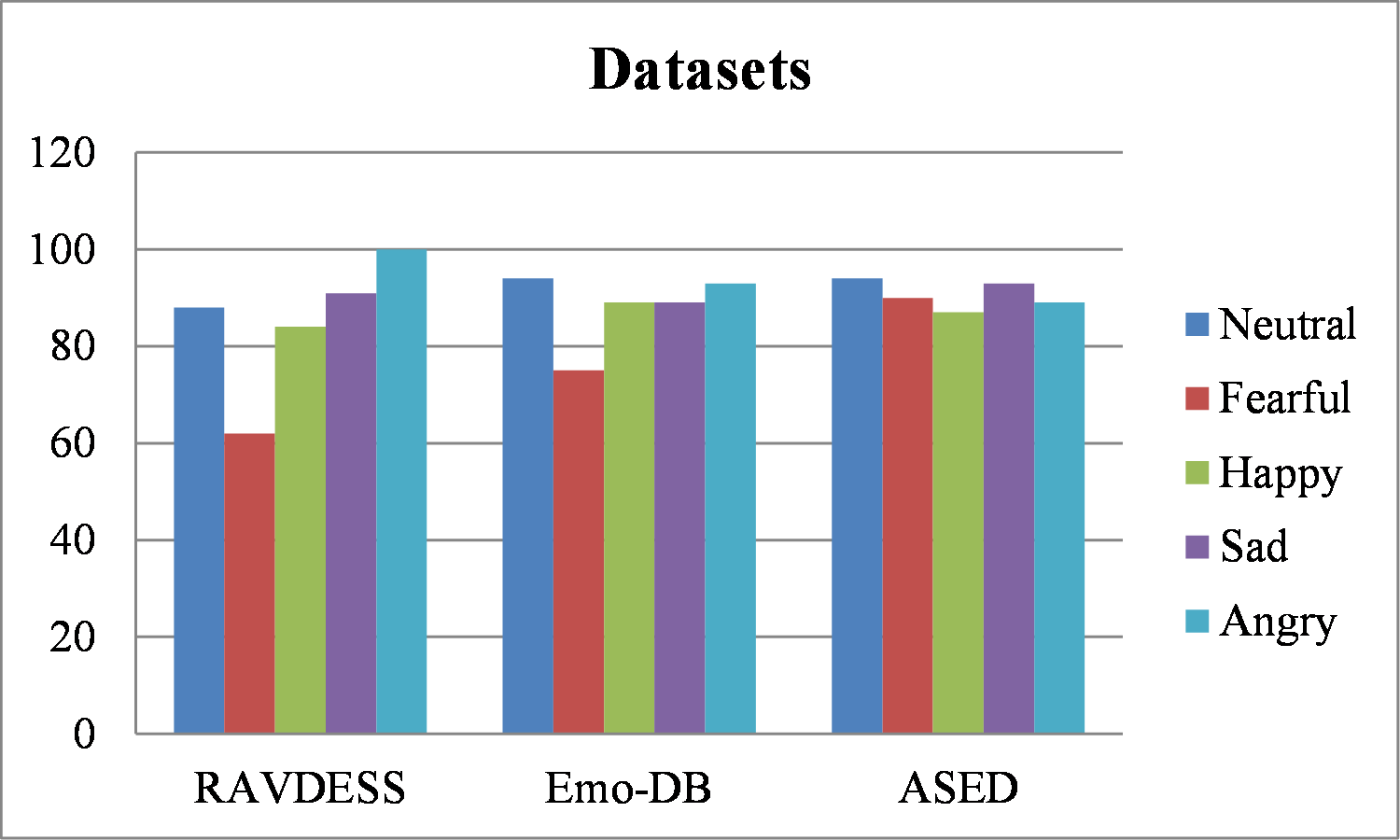}
			\caption{ Performance of VGGb on the RAVDESS, Emo-DB and ASED databases.}
			\label{performance-three-datasets}
		\end{figure}
		\FloatBarrier

Table \ref{vgg-accuracy-different-datasets} shows the results. Performance on ASED is the best (90.73\%) but all accuracy figures are within a range of 4.2\%. From this we can conclude that SER can be successfully performed on Amharic speech, and that the accuracy which can be attained is similar to that for English and German. The training time is longer (00:31:18) relative to EMO-DB (00:06:02) and RAVDESS (00:12:33) but this is because ASED is more than double the size of the other datasets. Fig. \ref{performance-three-datasets} shows the model's performance across the three datasets and across the five emotions. We can see that the fearful and sad classes are more clearly distinguished in ASED than in the other datasets. Conversely, the angry class is not as clearly distinguished as in the others.

\section{CONCLUSIONS}
In this paper, we first collected what we believe to be the very first SER dataset for the Amharic language, working with five emotions, neutral, fearful, happy, sad and angry. We then conducted three experiments, based on a four-layer version of VGG which we call VGGb. Experiment 1 was to determine whether Mel-spectrogram features or MFCC features were most suitable for SER in Amharic and to establish how robust this difference was relative to sentence choice, dialect and speaker group. Experiment 1.1 first compared performance on the whole dataset with standard cross-validation. Experiments 1.2-1.4 then tested the results using a form of cross-validation relative to sentences, dialects and speaker groups. The results showed that MFCC features were superior to Mel-spectrogram for Amharic SER and that this result was robust relative to the variations made.
   
In the second experiment, working with MFCC features and the ASED data, we compared the performance of four different models when applied to the SER task, Alex-Net, ResNet50, LSTM, and VGGb. While ResNet50 was the best (91.13\%), VGGb was very close (90.73\%) and was considerably faster (training time for ResNet50 08:44:15, for VGGb 00:31:18).

The third experiment compared the performance of VGGb on three datasets in different languages, RAVDESS (English), EMO-DB (German) and ASED (Amharic). The accuracies in each case were similar, suggesting that VGGb is equally applicable to the three languages.

Future work on ASED will include enlarging the scale of the database, using further elicitation techniques and adopting a dimensional emotion annotation strategy. We also plan to build a better SER model for Amharic and to study the extent to which emotion recognition is dependent on language.

\section*{Acknowledgments}
    This work was supported by The National Key Research and Development Program of China under grant 2020YFC1521503.

\bibliographystyle{abbrv}

\end{document}